%% file: eccv2022submission.tex
\documentclass[runningheads]{llncs}
\usepackage{graphicx}

\usepackage{tikz}
\usepackage{comment}
\usepackage{amsmath,amssymb} 
\usepackage{color}
\usepackage{xcolor}
\colorlet{Green}{green!80!black}

\usepackage[accsupp]{axessibility}  

\usepackage[pagebackref,breaklinks,colorlinks]{hyperref}

\usepackage{dsfont}

\usepackage{xspace}
\usepackage{pifont}
\usepackage{tabularx}
\newcolumntype{Y}{>{\centering\arraybackslash}X}
\usepackage{array,multirow}
\usepackage{wrapfig,lipsum,booktabs}
\newcommand{\cmark}{\ding{51}}
\newcommand{\xmark}{\ding{55}}
\usepackage{stmaryrd}
\usepackage{caption}
\usepackage{subcaption}
\usepackage{siunitx}
\usepackage{microtype}
\usepackage[subtle]{savetrees}
\usepackage{algorithm}
\usepackage{algpseudocode}
\usepackage{float}

\usepackage{orcidlink}

\usepackage[capitalize]{cleveref}
\crefname{section}{Sec.}{Secs.}
\Crefname{section}{Section}{Sections}
\Crefname{table}{Table}{Tables}
\crefname{table}{Tab.}{Tabs.}

\newcommand{\AP}{\text{AP}}

\newcommand{\hap}{\mathcal{H}\text{-AP}}
\newcommand{\hrank}{\mathcal{H}\text{-rank}}

\newcommand{\HAPPIER}{HAPPIER\xspace}

\newcommand{\lclust}{\mathcal{L}_\text{clust.}}

\newcommand{\lhap}{\mathcal{L}_{\hap}}
\newcommand{\lhaps}{\lhap^s}

\DeclareMathOperator{\rank}{rank}
\DeclareMathOperator{\rel}{rel}

\newcommand{\ie}{\emph{i.e.}\xspace}
\newcommand{\eg}{\emph{e.g.}\xspace}
\newcommand{\vs}{\emph{vs.}\xspace}
\newcommand{\etc}{\emph{etc.}\xspace}

\newcommand{\raone}{\scriptsize{R@1}}

\newcommand{\sap}{\scriptsize{AP}}

\usepackage{todonotes}

\begin{document}
\pagestyle{headings}
\mainmatter
\def\ECCVSubNumber{2061}  

\title{Hierarchical Average Precision Training for \\Pertinent Image Retrieval} 

\titlerunning{\HAPPIER}
%
\author{
Elias Ramzi\inst{1,2}\orcidlink{0000-0002-0131-2458} , Nicolas Audebert\inst{1}\orcidlink{0000-0001-6486-3102}, Nicolas Thome\inst{1,3}\orcidlink{0000-0003-4871-3045}, Clément Rambour\inst{1}\orcidlink{0000-0002-9899-3201}
\\\and Xavier Bitot\inst{2}
}
\authorrunning{E. Ramzi \textit{et al.}}
%
\institute{CEDRIC, Conservatoire National des Arts et Métiers, Paris, France \email{\{elias.ramzi,nicolas.audebert,nicolas.thome,clement.rambour\}@cnam.fr}\\
\and Coexya, Paris, France \\
\email{xavier.bitot@coexya.eu}\\\and Sorbonne Université, CNRS, ISIR, F-75005 Paris, France}
\maketitle

\setcounter{footnote}{0}

\begin{abstract}
   
   Image Retrieval is commonly evaluated with Average Precision (AP) or Recall@k. Yet, those metrics, are limited to binary labels and do not take into account errors' severity. This paper introduces a new hierarchical AP training method for pertinent image retrieval (HAPPIER). HAPPIER is based on a new $\hap$ metric, which leverages a concept hierarchy to refine AP by integrating errors' importance and better evaluate rankings. To train deep models with $\hap$, we carefully study the problem's structure and design a smooth lower bound surrogate combined with a clustering loss that ensures consistent ordering. Extensive experiments on 6 datasets show that HAPPIER significantly outperforms state-of-the-art methods for hierarchical retrieval, while being on par with the latest approaches when evaluating fine-grained ranking performances. Finally, we show that HAPPIER leads to better organization of the embedding space, and prevents most severe failure cases of non-hierarchical methods. Our code is publicly available at \url{https://github.com/elias-ramzi/HAPPIER}.
   
\end{abstract}

\input{files/introduction}

\input{files/related_works}

\input{files/method}

\input{files/experiments}

\input{files/conclusion}

\clearpage
%
%
\bibliographystyle{splncs04}
\bibliography{egbib}

\clearpage

\input{supplementary_file}

\end{document}

%% file: files/introduction.tex
\section{Introduction}\label{sec:introduction}

Image Retrieval (IR) consists in ranking images with respect to a query by decreasing order of visual similarity. IR methods are commonly evaluated using Recall@k (R@k) or Average Precision (AP). Because those metrics are non-differentiable, a rich literature exists on finding adequate surrogate loss functions to optimize them with deep learning, with tuple-wise losses~\cite{DBLP:conf/eccv/RadenovicTC16,NIPS2016_6b180037,wu2017sampling,wang2019multi,wang2020cross}, proxy based losses~\cite{zhai2018classification,wang2018cosface,deng2019arcface,teh2020proxynca++} and direct AP optimization methods~\cite{cakir2019deep,revaud2019learning,blackbox,blackboxap,smoothap,ramzi2021robust}.

\begin{figure*}[t]
    \centering
    \includegraphics[width=\textwidth]{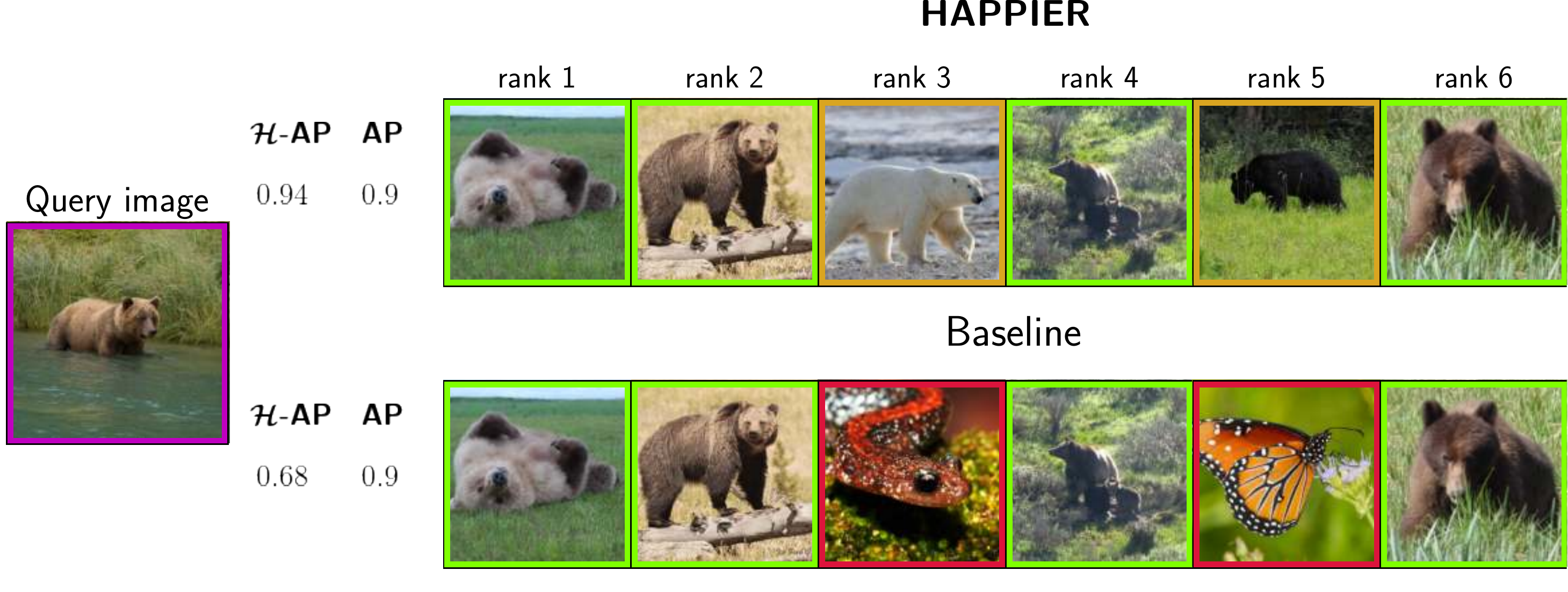}
    \caption{\textbf{Proposed \HAPPIER framework for pertinent image retrieval}. Standard ranking metrics based on binary labels, \eg Average Precision (AP), assign the same score to the  bottom and top row rankings ($0.9$). We introduce the $\hap$ metric based on non-binary labels, that takes into account mistakes' severity. $\hap$ assigns a smaller score to the bottom row ($0.68$) than the top one ($0.94$). \HAPPIER maximizes $\hap$ during training and thus explicitly supports to learn rankings similar to the top one, in contrast to binary ranking losses.
    }
    \label{fig:figure_intro}
\end{figure*}

These metrics are only defined for binary $(\oplus/\ominus)$ labels, which we denote as \emph{fine-grained labels}: an image is negative as soon as it is not strictly similar to the query. Binary metrics are by design unable to take into account the severity of the mistakes in a ranking. On~\cref{fig:figure_intro}, some negative instances are ``less negative'' than others, \eg given the ``Brown Bear'' query, ``Polar bear'' is more relevant than ``Butterfly''. However, \textcolor{black}{AP is $0.9$ for both} the top and bottom rankings. Consequently, training on binary metrics (\eg AP or R@k) develops no incentive to produce ranking such as the top row, and often produces rankings similar to the bottom one.
To address this problem, we introduce the \HAPPIER method dedicated to Hierarchical Average Precision training for Pertinent ImagE Retrieval. \HAPPIER provides a smooth training objective, amenable to gradient descent, which explicitly takes into account the severity of mistakes when evaluating rankings.

\definecolor{amethyst}{rgb}{0.6, 0.4, 0.8}

Our first contribution is to define a new Hierarchical AP metric ($\hap$) that leverages the hierarchical tree between concepts and enables a fine weighting between errors in rankings. As shown in~\cref{fig:figure_intro}, $\hap$ assigns a larger score ($0.94$) to the top ranking than to the bottom one ($0.68$). We show that $\hap$ provides a consistent generalization of AP for the non-binary setting. We also introduce our $\text{\HAPPIER}_\text{F}$ variant, giving more weights to fine-grained levels of the hierarchy. Since $\hap$, like AP, is a non-differentiable metric, our second contribution is to use \HAPPIER to directly optimize $\hap$ by gradient descent. We carefully design a smooth surrogate loss for $\hap$ that has strong theoretical guarantees and is an upper bound of the true loss. We then define an additional clustering loss to support having a consistency between partial and global rankings.

We validate \HAPPIER on six IR datasets, including three standard datasets (Stanford Online Products~\cite{oh2016deep} and iNaturalist-base/full~\cite{van2018inaturalist}), and three recent hierarchical datasets (DyML~\cite{sun2021dynamic}). We show that, when evaluating on hierarchical metrics (\eg $\hap$), \HAPPIER outperforms state-of-the-art methods for fine-grained ranking~\cite{wu2017sampling,zhai2018classification,teh2020proxynca++,ramzi2021robust}, the baselines and the latest hierarchical method of~\cite{sun2021dynamic}, and only slightly under-performs \vs state-of-the-art IR methods at the fine-grained level (\eg AP, R@1).
$\text{\HAPPIER}_\text{F}$ performs on par on fine-grained metrics while still outperforming fine-grained methods on hierarchical metrics.

%% file: files/related_works.tex
\section{Related work}\label{sec:related_works}

\subsection{Image Retrieval and ranking}
The Image Retrieval community has designed several families of methods to optimize metrics such as AP and R@k. Methods that relies on tuplet-wise losses, like pair losses~\cite{hadsell2006dimensionality,10.1007/978-3-319-46448-0_1}, triplet losses~\cite{wu2017sampling}, or larger tuplets~\cite{NIPS2016_6b180037,law2017learning,wang2019multi} learn comparison relations between instances. Methods using proxies have been introduced to lower the computational complexity of tuplet based training~\cite{movshovitz2017no,zhai2018classification,wang2018cosface,deng2019arcface,teh2020proxynca++}: they learn jointly a deep model and weight matrix that represent proxies using a cross-entropy based loss. Proxies are approximations of the original data points that should belong to their neighbourhood. Finally, there also has been large amounts of work dedicated to the direct optimization of the AP during training by introducing differentiable surrogates~\cite{cakir2019deep,revaud2019learning,blackbox,blackboxap,smoothap,ramzi2021robust}, so that models are optimized on the same metric they are evaluated on. However, nearly all of these methods only consider binary labels: two instances are either the same (positive) or different (negative), leading to poor performance when multiple levels of hierarchy are considered.

\subsection{Hierarchical predictions and metrics}

There has been a recent regain of interest in Hierarchical Classification~\cite{dhall2020hierarchical,bertinetto2020making,chang2021your} with the introductions of methods based either on a hierarchical softmax function or on multiple classifiers. It is considered that learning from hierarchical relations between labels leads to more robust models that make ``better mistakes" \cite{bertinetto2020making}. Yet, hierarchical classification means that labels are known in advance and are identical in the train and test sets. This is called a \emph{closed set} setting. However, Hierarchical Image Retrieval does not fall into this framework. Standard IR protocols consider the \emph{open set} paradigm to better evaluate the generalization abilities of learned models: the retrieval task at test time pertains to labels that were not present in the train set, making classification poorly suited to IR.

Meanwhile, the broader Information Retrieval community has been using datasets where documents can be more or less relevant depending on the query and the user making the request~\cite{hjorland2010foundation,graded_relevance}. Instead of the mere positive/negative dichotomy, each instance has a continuous score quantifying its relevance to the query. To quantify the quality of their retrieval engine, Information Retrieval researchers have long used ranking based metrics, such as the NDCG~\cite{jarvelin2002cumulated,croft2010search}, that penalize mistakes differently based on whether they occur at the top or the bottom of the ranking and whether wrong documents still have some marginal relevance or not. Average Precision is also used as a retrieval metric~\cite{jarvelin2017ir} and has even been given probabilistic interpretations based on how users interact with the system~\cite{pap}.
Several works have investigated how to optimize those metrics during the training of neural networks, \eg using pairwise losses~\cite{ranknet} and later using smooth surrogates of the NDCG in LambdaRank~\cite{lambdarank}, SoftRank~\cite{softrank}, ApproxNDCG~\cite{Qin2009AGA} and Learning-To-Rank~\cite{bruch2019revisiting}. These works however focused on NDCG, the most popular metric for information retrieval, and are without any theoretical guarantees: the surrogates are approximations of the NDCG but not \emph{lower bounds}, \ie their maximization does not imply improved performances during inference.

An additional drawback of this literature is that NDCG does not relate easily to average precision~\cite{dupret_2011}, which is the most common metric in image retrieval. Fortunately, there have been some works done to extend AP in a graded setting where relevance between instances is not binary~\cite{robertson2010extending,pap}. The graded Average Precision from~\cite{robertson2010extending} is the closest to our goal as it leverages SoftRank for direct optimization on non-binary relevance judgements, although there are significant shortcomings. There is no guarantee that the SoftRank surrogate actually minimizes the graded AP, it requires to annotate datasets with pairwise relevances which is unpractical for large scale settings and was only applied to small-scale corpora of a few thousands documents, compared to the hundred thousands of images in IR.

Recently, the authors of~\cite{sun2021dynamic} introduced three new hierarchical benchmarks datasets for image retrieval, in addition to a novel hierarchical loss CSL. CSL extends proxy-based triplet losses to the hierarchical setting and tries to structure the embedding space in a hierarchical manner. However, this method faces the same limitation as the usual triplet losses: minimizing CSL does not explicitly optimize a well-behaved hierarchical evaluation metric, \eg $\hap$. We show experimentally that our method \HAPPIER significantly outperforms CSL~\cite{sun2021dynamic} both on hierarchical metrics and AP-level evaluations.

%% file: files/method.tex
\section{HAPPIER Model}
\label{sec:methods}

We detail \HAPPIER our Hierarchical Average Precision training method for Pertinent ImagE Retrieval. We first introduce the Hierarchical Average Precision, $\hap$ in~\cref{sec:hierarchical_metric}, that leverages a hierarchical tree (\cref{fig:figure_method}a) of labels. It is based on the hierarchical rank, $\hrank$, and evaluates rankings so that more relevant instances are ranked before less relevant ones (\cref{fig:figure_method}b). We then show how to directly optimize $\hap$ by stochastic gradient descent (SGD) using \HAPPIER in~\cref{sec:direct_optimization}. Our training objective combines a carefully designed smooth upper bound surrogate loss for $\lhap=1-\hap$ and a clustering loss $\lclust$ that supports consistent rankings.

\definecolor{amethyst}{rgb}{0.6, 0.4, 0.8}
\begin{figure}[t]
    \centering
    \includegraphics[width=1\textwidth]{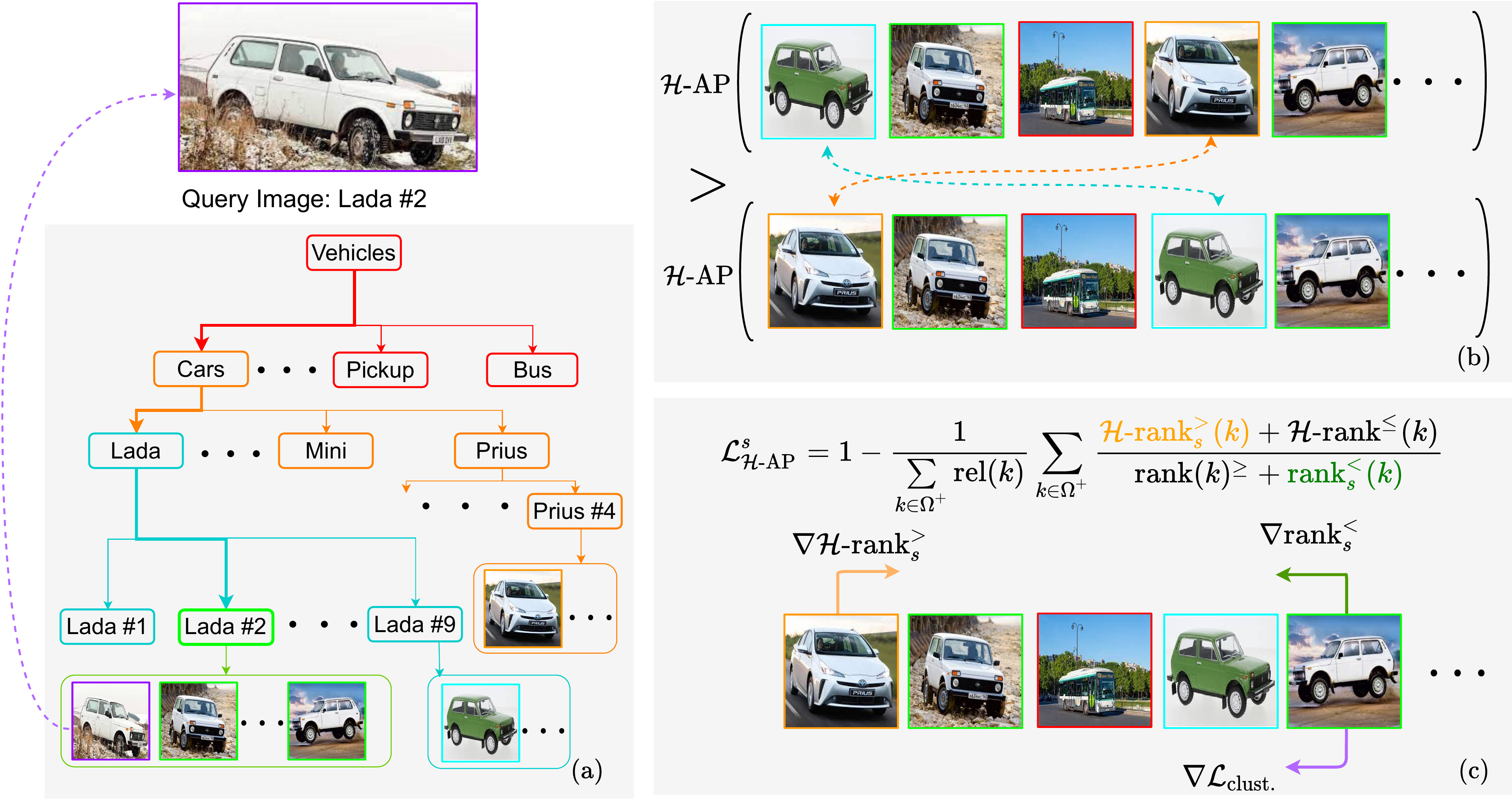}
    \caption{
    HAPPIER leverages a hierarchical tree representing the semantic similarities between concepts in (a) to introduce a new hierarchical metric, $\hap$ in \cref{eq:def_hap}, see (b). $\hap$ exploits the hierarchy to weight rankings' inversion:  given the \textcolor{amethyst}{query image} of a ``\textcolor{Green}{Lada \#2}'', $\hap$ penalizes an inversion with a ``\textcolor{cyan}{Lada \#9}'' less than with a ``\textcolor{orange}{Prius \#4}''. To directly train models with $\hap$, we carefully study the structure of the problem and introduce the $\lhaps$ loss in~\cref{eq:rewrite_hap}, which provides a smooth upper bound of $\lhap$, see (c). We also train HAPPIER with the $\lclust$ loss in~\cref{eq:cluster_loss} to enforce the partial ordering in stochastic optimization to mach the global ones.
    }
    \label{fig:figure_method}
\end{figure}

\textbf{Context} Let us consider a retrieval set $\Omega=\left\{\boldsymbol{x_j}\right\}_{j \in \llbracket 1;N\rrbracket}$ composed of $N$ instances. For a query\footnote{For the sake of readability, our notations are given for a single query. During training, \HAPPIER optimizes our hierarchical retrieval objective by averaging several queries.} $\boldsymbol{q} \in \Omega$, we aim to order all $x_j \in \Omega$ so that more relevant (\ie similar) instances are ranked before less relevant instances. 

In our hierarchical setting, the relevance of an instance $\boldsymbol{x_j}$ is non-binary. We assume that we have access to a hierarchical tree defining semantic similarities between concepts as in~\cref{fig:figure_method}a. For a query $\boldsymbol{q}$, we leverage this knowledge to partition the set of retrieved instances into $L+1$ disjoint subsets $\left\{\Omega^{(l)}\right\}_{l \in \llbracket 0;L\rrbracket}$. $\Omega^{(L)}$ is the subset of the most similar instances to the query (\ie fine-grained level): for $L=3$ and a ``Lada \#2''  query, $\Omega^{(3)}$ are the images of the same ``Lada \#2'' (green), see~\cref{fig:figure_method}. The set $\Omega^{(l)}$ for $l<L$ contains instances with smaller relevance with respect to the query: $\Omega^{(2)}$ in \cref{fig:figure_method} is the set of ``Lada'' that are not ``Lada \#2'' (blue) and $\Omega^{(1)}$ is the set of ``Cars'' that are not ``Lada'' (orange). We also define $\Omega^- := \Omega^{(0)}$ as the set of negative instances, \ie the set of vehicles that are not ``Cars'' (in red) in~\cref{fig:figure_method} and $\Omega^+ = \bigcup_{l=1}^L \Omega^{(l)}$.
Each instance $k$ of $\Omega^{(l)}$ is thus associated a value through the \emph{relevance function} denoted as $\boldsymbol{\rel(k)}$~\cite{hjorland2010foundation}.

To rank the instances $x_j \in \Omega$ with respect to the query $\boldsymbol{q}$, we compute cosine similarities in an embedding space. More precisely, we extract embedding vectors using a deep neural network $\boldsymbol{f}$ parameterized by $\boldsymbol{\theta}$, $v_{j} = f_\theta(x_j)$, and compute the cosine similarity between the query and every image $s_{j} = f_\theta(q)^T v_{j}$. Images are then ranked by decreasing cosine similarity score. We learn the parameters $\boldsymbol{\theta}$ of the network with \HAPPIER, our framework to directly minimize ${\mathcal{L}_{\hap}(\theta)=1-\hap(\theta)}$. This enforces a ranking where the instances with the highest cosine similarity scores belong to $\Omega^{(L)}$, then $\Omega^{(L-1)}$ \etc and the items with the lowest cosine similarity belong to $\Omega^-$.

\subsection{Hierarchical Average Precision}\label{sec:hierarchical_metric}
\label{sec:hap}

Average Precision (AP) is the most common metric in Image Retrieval. AP evaluates a ranking in a binary setting: for a given query, each instance is either positive or negative. It is computed as the average of precision at each rank $n$ over the positive set $\AP = \frac{1}{|\Omega^+|} \sum_{n=1}^{N} \operatorname{Prec}(n)$. Previous works have written the AP using the ranking operator~\cite{smoothap} as in~\cref{eq:ap_definition}. The rank for an instance $k$ is written as a sum of Heaviside (step) function $H$~\cite{Qin2009AGA}: this counts the number of instances $j$ ranked before $k$, \ie that have a higher cosine similarity ($s_j > s_k$). $\rank^+$ is the rank among the positive instances, \ie restricted to $\Omega^+$.

\begin{equation}
    \label{eq:ap_definition}
    \text{AP} = \frac{1}{|\Omega^+|} \sum_{k\in\Omega^+} \frac{\rank^+(k)}{\rank(k)}, \; \text{with }
    \begin{cases}
      \rank(k) = 1 + \sum_{j\in\Omega} H(s_j - s_k) \\
      \rank^+(k) = 1 + \sum_{j\in\Omega^+} H(s_j - s_k)
    \end{cases}
\end{equation}

\subsubsection{Extending AP to hierarchical image retrieval}

We propose an extension of AP that leverages non-binary labels. To do so, we extend the concept of $\rank^+$ to the hierarchical case with the concept of hierarchical rank, $\hrank$:

\begin{figure}[t]
    \centering
    \includegraphics[width=0.8\textwidth]{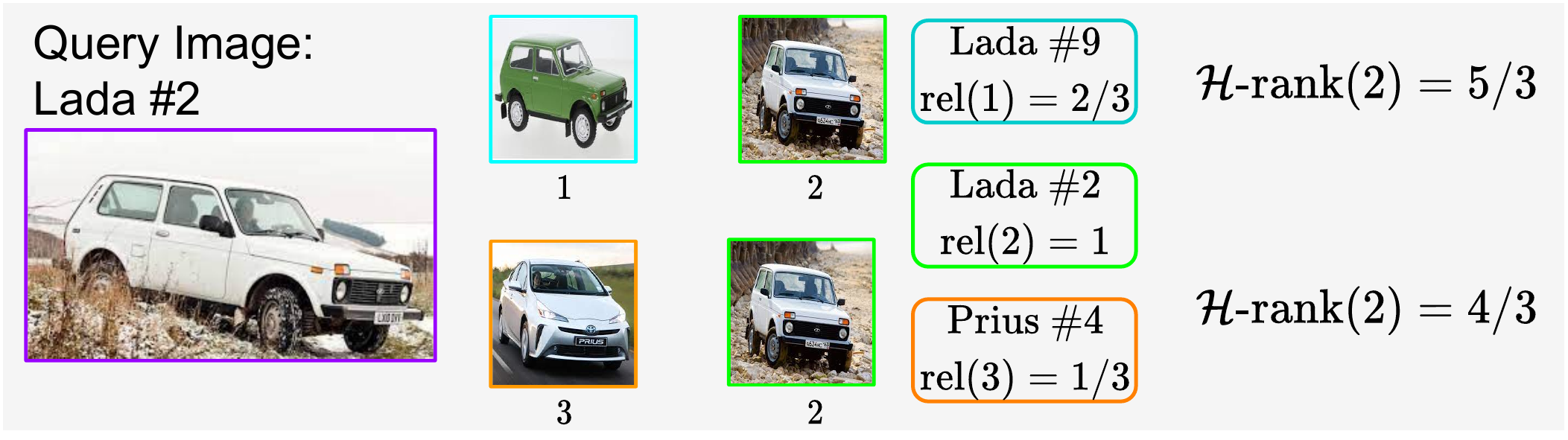}
    \caption{
    Given a ``\textcolor{Green}{Lada \#2}'' \textcolor{amethyst}{query}, the top inversion is less severe than the bottom one. Indeed on the top row instance $1$ is semantically closer to the query -- as it is a ``\textcolor{cyan}{Lada}''-- than instance $3$ on the bottom row. Indeed instance $3$'s closest common ancestor with the query, ``\textcolor{orange}{Cars}'', is farther in the hierarchical tree (see~\cref{fig:figure_method}a). Because of that $\hrank(2)$ is greater on the top row ($5/3$) than on the bottom row ($4/3$), leading to a greater $\hap$ in \cref{fig:figure_method}b for the top row.
    }
    \label{fig:figure_hrank}
\end{figure}

\begin{equation}
    \hrank(k) = \rel(k) + \sum_{j\in\Omega^+} \min(\rel(k), \rel(j))\cdot H(s_j-s_k) ~.
    \label{eq:hierarchical_rank}
\end{equation}

Intuitively, $\min(\rel(k), \rel(j))$ corresponds to seeking the closest ancestor shared by instance $k$ and $j$ with the query in the hierarchical tree. As illustrated in~\cref{fig:figure_hrank}, $\hrank$ induces a smoother penalization for instances that do not share the same fine-grained label as the query but still share some coarser semantics, which is not the case for $\rank^+$.

From $\hrank$ in~\cref{eq:hierarchical_rank} we define the Hierarchical Average Precision, $\hap$:
\begin{equation}
\label{eq:def_hap}
    \hap = \frac{1}{\sum_{k\in\Omega^+}\rel(k)} \sum_{k\in\Omega^+} \frac{\hrank(k)}{\rank(k)}
\end{equation}
\cref{eq:def_hap} extends the AP to non-binary labels. We replace $\rank^+$ by our hierarchical rank $\hrank$ and the normalization term $|\Omega^+|$ is replaced by $\sum_{k\in\Omega^+}\rel(k)$, which both represent the ``sum of positives'', see more details in supplementary A.2.

$\hap$ extends the desirable properties of the AP. It evaluates the quality of a ranking by: i) penalizing inversions of instances that are not ranked in decreasing order of relevances with respect to the query, ii) giving stronger emphasis to inversions that occur at the top of the ranking. Finally, we can observe that, by this definition, $\hap$ is equal to the AP in the binary setting ($L=1$). This makes $\hap$ a \emph{consistent generalization} of AP (details in supplementary A.2).

\subsubsection{Relevance function design}\label{seq:relevance}

The relevance $\rel(k)$ defines how ``similar'' an instance $k\in\Omega^{(l)}$ is to the query $q$. While $\rel(k)$ might be given as input in Information Retrieval datasets \cite{DBLP:journals/corr/QinL13,chapelle2011yahoo}, we need to define it based on the hierarchical tree in our case. We want to enforce the constraint that the relevance decreases when going up the tree, \ie $\rel(k)>\rel(k')$ for $k\in\Omega^{(l)}$, $k'\in\Omega^{(l')}$ and $l>l'$. To do so, we assign a total weight of $(l/L)^\alpha$ to each semantic level $l$, where $\alpha\in\mathbb{R}^+$ controls the decrease rate of similarity in the tree. For example for $L=3$ and $\alpha=1$, the total weights for each level are $1$, $\frac{2}{3}$, $\frac{1}{3}$ and $0$. The instance relevance $\rel(k)$ is normalized by the cardinal of $\Omega^{(l)}$:
\begin{equation}\label{eq:hierarchy_relevance}
    \rel(k) = \frac{(l/L)^\alpha}{|\Omega^{(l)}|} \; \text{if } k \in \Omega^{(l)}
\end{equation}

\medbreak
Other definitions fulfilling the decreasing similarity behaviour in the tree are possible.
An interesting option for the relevance enables to recover a weighted sum of AP, denoted as $\sum w\AP:=\sum_{l=1}^L w_l \cdot \AP^{(l)}$ (supplementary A.2), \ie the weighted sum of AP is a particular case of $\hap$.

We set $\alpha=1$ in~\cref{eq:hierarchy_relevance} for the $\hap$ metric and in our main experiments. Setting $\alpha$ to larger values supports better performances on fine-grained levels as their relevances will \textcolor{black}{relatively} increase. This variant is denoted $\text{\HAPPIER}_{\text{F}}$ and discussed in \cref{sec:experiments}.

\subsection{Direct optimization of $\hap$}\label{sec:direct_optimization}

$\hap$ in \cref{eq:def_hap} involves the computation of $\hrank$ and rank, which are non-differentiable due to the summing of Heaviside step functions. We thus introduce a smooth approximation of $\hap$ to obtain a surrogate loss amenable to gradient descent, which fulfils theoretical guarantees for proper optimization.

\medbreak
\medbreak
\noindent\textbf{Re-writing $\hap$} In order to design our surrogate loss for $\lhap=1-\hap$, we decompose $\hrank$ and $\rank$ into two quantities. Denoting $\hrank^>(k)$ (resp. $\hrank^\leq(k)$) as the restriction of $\hrank$ to instances of strictly higher relevances (resp. lower or equal), we can see that $\hrank(k) = \hrank^>(k) + \hrank^\leq(k)$. The rank can be decomposed in a similar fashion: $\rank(k) = \rank^\geq(k) + \rank^<(k)$ where $<$ (resp. $\geq$) denotes the restriction to instances of strictly lower relevances (resp. higher or equal). The $\lhap$ can be rewritten as follow:

\begin{equation}\label{eq:rewrite_hap}
    \lhap = 1 - \frac{1}{\sum_{k\in\Omega^+}\rel(k)} \sum_{k\in\Omega^+} \frac{\hrank^>(k) + \hrank^\leq(k)}{\rank^\geq(k) + \rank^<(k)}~.
\end{equation}

We choose to optimize over $\hrank^>$ and $\rank^<$ in~\cref{eq:rewrite_hap}. We \textcolor{black}{maximize} $\hrank^>$ to enforce that the $k$\textsuperscript{th} instance must decrease in cosine similarity score if it is ranked before another instance of higher relevance ($\nabla \hrank^>$ in~\cref{fig:figure_method} enforces the blue instance to be ranked after the green one as it is less relevant to the query). We \textcolor{black}{minimize} $\rank^<$ to encourage the $k$\textsuperscript{th} instance to increase in cosine similarity score if it is ranked after one or more instances of lower relevance ($\nabla \rank^<$ in~\cref{fig:figure_method} enforces that the last green instance moves before less relevant instances). Optimizing both those terms leads to a decrease in $\lhap$. On the other hand, we purposely do not optimize the two remaining $\hrank^\leq(k)$ and $\rank^\geq(k)$ terms, since this could harm training performances as explained in supplementary~A.3.

\medbreak
\noindent\textbf{Upper bound of $\lhap$} Based on the previous analysis, we now design our surrogate loss $\lhaps$ by introducing a smooth approximation of $\rank^<$ and $\hrank^>(k)$. An important sought property of $\lhaps$ is that it is an upper bound of $\lhap$. To this end, we approximate $\hrank^>(k)$ with a piece-wise linear function that is a lower bound of the Heaviside function. $\rank^<$ is approximated with a smooth upper bound of the Heaviside that combines a piece-wise sigmoid function and an affine function, which has been shown to make the training more robust thanks to the induced implicit margins between positives and negatives~\cite{blackboxap,smoothap,ramzi2021robust}. More details are given in supplementary A.3 on those surrogates.

\subsubsection{Clustering constraint in \HAPPIER} Positives only need to have a greater cosine similarity with the query than negatives in order to be correctly ranked. Yet, we cannot optimize the ranking on the entire datasets -- and thus the true $\lhap$ -- because of the batch-wise estimation performed in stochastic gradient descent. To mitigate this issue, we take inspiration from clustering methods~\cite{zhai2018classification,teh2020proxynca++} to define the following objective in order to group closely the embeddings of instances that share the same fine-grained label:

\begin{equation}\label{eq:cluster_loss}
    \lclust(\theta) = - \log\left( \frac{\exp(\frac{v_y^T p_y}{\sigma})}{\sum_{{p_z}\in\mathcal{Z}} \exp(\frac{v_y^T p_z}{\sigma})} \right),
\end{equation}
\noindent where $p_y$ is the normalized proxy corresponding to the fine-grained class of the embedding $v_y$, $\mathcal{Z}$ is the set of proxies, and $\sigma$ is a temperature scaling parameter.
In~\cref{fig:figure_method}, $\nabla\lclust$ further clusters ``Lada \#2'' instances. $\lclust$ induces a reference shared across batches and thus enforces that the partial ordering in-between batches is consistent with the global ordering over the entire retrieval set.

\medbreak
Our resulting final objective is a linear combination of both our losses, with a weight factor $\lambda\in[0,1]$ that balances the two terms:

\begin{equation*}
    \mathcal{L}_{\text{HAPPIER}}(\theta) = (1-\lambda)\cdot\lhaps(\theta) + \lambda \cdot \lclust(\theta) ~.
\end{equation*}

%% file: files/experiments.tex
\section{Experiments}\label{sec:experiments}

\subsection{Experimental setup}\label{sec:experimental_setup}

\noindent\textbf{Datasets} We use the standard benchmark Stanford Online Products~\cite{oh2016deep} (SOP) with two levels of hierarchy ($L=2$), and iNaturalist-2018~\cite{van2018inaturalist} with the standard
~splits from~\cite{smoothap} in two settings: i) iNat-base with two levels of hierarchy ($L=2$) ii) iNat-full with the full biological taxonomy composed of 7 levels ($L=7$). We also evaluate on the recent dynamic metric learning (DyML) datasets (DyML-V, DyML-A, DyML-P) introduced in~\cite{sun2021dynamic} for the task of hierarchical image retrieval, each with 3 semantic levels ($L=3$).

\medbreak
\noindent\textbf{Implementation details} Our base model is a ResNet-50 pretrained on ImageNet for SOP and iNat-base/full, and a ResNet-34 randomly initialized on DyML-V\&A and pretrained on ImageNet on DyML-P, following~\cite{sun2021dynamic}.
Unless specified otherwise, all reported results are obtained with $\alpha=1$ in~\cref{eq:hierarchy_relevance} and $\lambda=0.1$ for $\mathcal{L}_{\text{\HAPPIER}}$. We study the impact of these parameters in~\cref{sec:happier_analysis}.

\medbreak
\noindent\textbf{Metrics} For SOP and iNat, we evaluate the models based on three hierarchical metrics: $\hap$ -- which we introduced in~\cref{eq:def_hap} -- the Average Set Intersection (ASI) and the Normalized Discounted Cumulative Gain (NDCG), defined in supplementary B.3. We also report the AP for each semantic level. For DyML, we follow the evaluation protocols of~\cite{sun2021dynamic} and compute AP, ASI and R@1 on each semantic scale before averaging them. We cannot compute $\hap$ or NDCG on those datasets as the hierarchical tree is not available on the test set. 

\medbreak
\noindent\textbf{Baselines}
We compare \HAPPIER to several recent image retrieval methods optimized at the fine-grained level, which represent strong baselines for IR when training with binary labels: Triplet SH ($\text{TL}_\text{SH}$) \cite{wu2017sampling}, NormSoftMax (NSM) \cite{zhai2018classification}, ProxyNCA++ (NCA++) \cite{teh2020proxynca++} and ROADMAP \cite{ramzi2021robust}. We also benchmark against hierarchical methods obtained by summing these fine-grained losses at different levels (denoted by $\Sigma$), and with respect to the recent hierarchical CSL loss~\cite{sun2021dynamic}.\\
Details on the experimental setup are given in supplementary B.

\subsection{Main Results}\label{sec:main_results}

\subsubsection{Hierachical results}
We first evaluate \HAPPIER on global hierarchical metrics.
On \cref{tab:main_sop_inat}, we notice that \HAPPIER significantly outperforms methods trained on the fine-grained level only, with a gain on $\hap$ over the best performing methods of +16.1pt on SOP, \textcolor{black}{+13pt}
~on iNat-base and \textcolor{black}{12.7pt}
~on iNat-full. \HAPPIER also exhibits significant gains compared to hierarchical methods. On $\hap$, \HAPPIER has important gains on all datasets (\eg +6.3pt on SOP, +4.2pt on iNat-base over the best competitor), but also on ASI and NDCG. This shows the strong generalization of the method on standard metrics. Compared to the recent CSL loss~\cite{sun2021dynamic}, we observe a consistent gain over all metrics and datasets, \eg +6pt on $\hap$, +8pt on ASI and +2.6pts on NDCG on SOP. This shows the benefits of optimizing a well-behaved hierarchical metric compared to an ad-hoc proxy method.

\begin{table*}[ht]
    \caption{Comparison of \HAPPIER on SOP and iNat-base/full when using hierarchical metrics. Best results in \textbf{bold}, second best \underline{underlined}.}
    \label{tab:main_sop_inat} 
    \centering
    \begin{tabularx}{\textwidth}{ l l YYY | YYY | YYY}
        \toprule
        & \multirow{2}{*}{Method}  & \multicolumn{3}{c|}{SOP} & \multicolumn{3}{c|}{iNat-base} & \multicolumn{3}{c}{iNat-full} \\
        \cmidrule{3-11}
        && \multirow{1}{*}{$\mathcal{H}$-AP} & ASI & \multirow{1}{*}{NDCG} & \multirow{1}{*}{$\mathcal{H}$-AP} & ASI & \multirow{1}{*}{NDCG} & \multirow{1}{*}{$\mathcal{H}$-AP} & ASI & \multirow{1}{*}{NDCG} \\
         \midrule
         \multirow{5}{*}{\rotatebox[origin=c]{90}{Fine}}
         & Triplet SH~\cite{wu2017sampling} & 42.2 & 22.4 & 78.8 & 39.5 & 63.7 & 91.5 & 36.1 & 59.2 & 89.8 \\
         & NSM~\cite{zhai2018classification} & 42.8 & 21.1 & 78.3 & 38.0 & 51.6 & 88.9 & 33.3 & 51.7 & 88.2 \\
         & NCA++~\cite{teh2020proxynca++} & 43.0 & 21.5 & 78.4 & 39.5 & 57.0 & 90.1 & 35.3 & 55.7 & 89.0  \\
         & \textcolor{black}{Smooth-AP~\cite{smoothap}} & 42.9 & 20.6 & 78.2 & 41.3 & 64.2 & 91.9 & 37.2 & 60.1 & 90.1 \\
         & ROADMAP~\cite{ramzi2021robust} & 43.3 & 19.1 & 77.9 & 40.3 & 61.0 & 91.2 & 34.7 & 59.6 & 89.5 \\
         \midrule
         \multirow{5}{*}{\rotatebox[origin=c]{90}{Hier.}}
         & $\Sigma\text{TL}_{\text{SH}}$~\cite{wu2017sampling} & \underline{53.1} & 53.3 & \underline{89.2} & 44.0 & 87.4 & 96.4 & 39.9 & \underline{85.5} & 92.0 \\ 
         & $\Sigma$NSM~\cite{zhai2018classification} & 50.4 & 49.7 & 87.0 & 47.9 & 75.8 & 94.4 & \underline{46.9} & 74.2 & \textbf{93.8}  \\
         & $\Sigma$NCA++~\cite{teh2020proxynca++} & 49.5 & 52.8 & 87.8 & 48.9 & 78.7 & 95.0 & 44.7 & 74.3 & 92.6 \\
         & CSL~\cite{sun2021dynamic} & 52.8 & \underline{57.9} & 88.1 & \underline{50.1} & \textbf{89.3} & \underline{96.7} & 45.1 & 84.9 & 93.0 \\ 
        \cmidrule{2-11}
         & \textbf{\HAPPIER} & \textbf{59.4} & \textbf{65.9} & \textbf{91.5} & \textbf{54.3} & \textbf{89.3} & \textbf{96.9} & \textbf{47.9} & \textbf{87.2} & \textbf{93.8} \\
        \bottomrule
    \end{tabularx}
\end{table*}

On~\cref{tab:main_dyml}, we evaluate \HAPPIER on the recent DyML benchmarks. \HAPPIER again shows significant gains in mAP and ASI compared to methods only trained on fine-grained labels, \eg +9pt in mAP and +10pt in ASI on DyML-V. \HAPPIER also outperforms other hierarchical baselines: +4.8pt mAP on DyML-V, +0.9 on DyML-A and +1.8 on DyML-P. In R@1, \HAPPIER performs on par with other methods on DyML-V and outperforms other hierarchical baselines by a large margin on DyML-P: 63.7 \vs 60.8 for $\Sigma$NSM. Interestingly, \HAPPIER also consistently outperforms CSL~\cite{sun2021dynamic} on its own datasets\footnote{CSL's score  on~\cref{tab:main_dyml} are above those reported in~\cite{sun2021dynamic}; personal discussions with the authors~\cite{sun2021dynamic} validate that our results are valid for CSL, see supplementary B.5.}.

\begin{table*}[ht]
    \caption{Performance comparison on Dynamic Metric Learning benchmarks~\cite{sun2021dynamic}.
    }
    \label{tab:main_dyml} 
    \centering
    \begin{tabularx}{\textwidth}{l l YYY | YYY | YYY }
        \toprule
         & \multirow{2}{*}{Method} & \multicolumn{3}{c|}{DyML-Vehicle} & \multicolumn{3}{c|}{DyML-Animal} & \multicolumn{3}{c}{DyML-Product}\\
        \cmidrule{3-11}
         && mAP & ASI & R@1 & mAP & ASI & R@1 & mAP & ASI & R@1 \\
         \midrule
         \multirow{4}{*}{\rotatebox[origin=c]{90}{Fine}}
         & $\text{TL}_{\text{SH}}$~\cite{wu2017sampling} & 26.1 & 38.6 & 84.0 & 37.5 & 46.3 & 66.3 & 36.32 & 46.1 & 59.6 \\
         & NSM~\cite{zhai2018classification} & 27.7 & 40.3 & 88.7 & 38.8 & 48.4 & \underline{69.6} & 35.6 & 46.0 & 57.4 \\
         & \textcolor{black}{Smooth-AP~\cite{smoothap}} & 27.1 & 39.5 & 83.8 & 37.7 & 45.4 & 63.6 & 36.1 & 45.5 & 55.0  \\
         & ROADMAP~\cite{ramzi2021robust} & 27.1 & 39.6 & 84.5 & 34.4 & 42.6 & 62.8 & 34.6 & 44.6 & \underline{62.5} \\
        \midrule
        \multirow{4}{*}{\rotatebox[origin=c]{90}{Hier.}}
         & $\Sigma\text{TL}_{\text{SH}}$~\cite{wu2017sampling} & 25.5 & 38.1 & 81.0 & 38.9 & 47.2 & 65.9 & \underline{36.9} & 46.3 & 58.5 \\
         & $\Sigma$NSM~\cite{zhai2018classification} & \underline{32.0} & \underline{45.7} & \textbf{89.4} & \underline{42.6} & \underline{50.6} & \textbf{70.0} & 36.8 & \underline{46.9} & 60.8 \\
        & CSL~\cite{sun2021dynamic} & 30.0 & 43.6 & 87.1 & 40.8 & 46.3 & 60.9 & 31.1 & 40.7 & 52.7 \\
         \cmidrule{2-11}
        & \textbf{\HAPPIER} & \textbf{37.0} & \textbf{49.8} & \underline{89.1} &  \textbf{43.8} & \textbf{50.8} & 68.9 & \textbf{38.0} & \textbf{47.9} & \textbf{63.7}\\
         \bottomrule
    \end{tabularx}
\end{table*}

\subsubsection{Detailed evaluation} \cref{tab:detail_sop_inat_base,tab:detail_inat_full} shows the different methods' performances on all semantic hierarchy levels. We evaluate HAPPIER and also $\text{\HAPPIER}_{\text{F}}$ ($\alpha >1$ for~\cref{eq:hierarchy_relevance} in~\cref{sec:hap}), with $\alpha=5$ on SOP and $\alpha=3$ on iNat-base/full. HAPPIER optimizes the overall hierarchical performances, while $\text{\HAPPIER}_{\text{F}}$ is meant to be optimal at the fine-grained level while still optimizing coarser levels.

\begin{table*}[ht]
    \setlength\tabcolsep{2pt}
    \caption{Comparison of \HAPPIER \vs methods trained only on fine-grained labels on SOP and iNat-base. Metrics are reported for both semantic levels.}
    \label{tab:detail_sop_inat_base} 
    \centering
    \begin{tabularx}{\textwidth}{ l l YYY | YYY}
        \toprule
         & & \multicolumn{3}{c|}{SOP} & \multicolumn{3}{c}{iNat-base} \\
        \cmidrule{3-8}
         & & \multicolumn{2}{c}{Fine} & \multicolumn{1}{c|}{Coarse} & \multicolumn{2}{c}{Fine} & \multicolumn{1}{c}{Coarse}  \\
         & Method & R@1 & AP & AP & R@1 & AP & AP \\
         \midrule
         \multirow{5}{*}{\rotatebox[origin=c]{90}{Fine}}
         & $\text{TL}_{\text{SH}}$~\cite{wu2017sampling} & 79.8 & 59.6 & 14.5 & 66.3 & 33.3 & 51.5 \\
         & NSM~\cite{zhai2018classification} & 81.3 & 61.3 & 13.4  & 70.2 & \underline{37.6} & 38.8 \\
         & NCA++~\cite{teh2020proxynca++} & 81.4 & 61.7 & 13.6 & 67.3 & 37.0 & 44.5 \\
         & \textcolor{black}{Smooth-AP~\cite{smoothap}} & 81.3 & 61.7 & 13.4 & 67.3 & 35.2 & 53.1 \\
         & ROADMAP~\cite{ramzi2021robust} & \textbf{82.2} & \textbf{62.5} & 12.9 & 69.3 & 35.1 & 50.4 \\
         
         \midrule
         \multirow{3}{*}{\rotatebox[origin=c]{90}{Hier.}}
         & CSL~\cite{sun2021dynamic} & 79.4 & 58.0 & \underline{45.0} & 62.9 & 30.2 & \underline{88.5} \\
         \cmidrule{2-8}
         & \textbf{\HAPPIER} & 81.0 & 60.4 & \textbf{58.4} & \underline{70.7} & 36.7 & \textbf{88.6} \\
         & \textbf{$\text{\HAPPIER}_{\text{F}}$} & \underline{81.8} & \underline{62.2} & 36.0 & \textbf{71.0} & \textbf{37.8} & 78.8 \\
        \bottomrule
    \end{tabularx}
\end{table*}

On~\cref{tab:detail_sop_inat_base}, we observe  that \HAPPIER gives the best performances at the coarse level, with a significant boost compared to fine-grained methods, \eg +43.9pt AP compared to the best non-hierarchical $\text{TL}_{\text{SH}}$~\cite{wu2017sampling} on SOP. \HAPPIER even outperforms the best fine-grained methods in R@1 on iNat-base, but is slightly below on SOP. $\text{\HAPPIER}_{\text{F}}$ performs on par with the best methods at the finest level on SOP, while further improving performances on iNat-base, and still significantly outperforms fine-grained methods at the coarse level.

The satisfactory behaviour and the two optimal regimes of \HAPPIER and $\text{\HAPPIER}_{\text{F}}$ are confirmed and even more pronounced on iNat-full (\cref{tab:detail_inat_full}): \HAPPIER gives the best results on coarser levels (from ``Order''), while being very close to the best results on finer ones. $\text{\HAPPIER}_{\text{F}}$ gives the best results at the finest levels, even outperforming very competitive fine-grained baselines.

Again, note that \HAPPIER outperforms CSL~\cite{sun2021dynamic} on all semantic levels and datasets on~\cref{tab:detail_sop_inat_base,tab:detail_inat_full}, \eg +5pt on the fine-grained AP (``Species'') and +3pt on the coarsest AP (``Kingdom'') on~\cref{tab:detail_inat_full}.

\begin{table*}[ht]
    \caption{Comparison of \HAPPIER \vs methods trained only on fine-grained labels on iNat-Full. Metrics are reported for all 7 semantic levels.}
    \label{tab:detail_inat_full} 
    \centering
    \begin{tabularx}{\textwidth}{l l YY Y Y Y Y Y Y }
        \toprule
        & \multirow{2}{*}{Method} & \multicolumn{2}{c}{\scriptsize{Species}} & \multicolumn{1}{c}{\scriptsize{Genus}} & \multicolumn{1}{c}{\scriptsize{Family}} & \multicolumn{1}{c}{\scriptsize{Order}} & \multicolumn{1}{c}{\scriptsize{Class}} & \multicolumn{1}{c}{\scriptsize{Phylum}} & \multicolumn{1}{c}{\scriptsize{Kingdom}} \\
         && \raone & \sap & \sap & \sap & \sap & \sap & \sap & \sap \\
         \midrule
         \multirow{5}{*}{\rotatebox[origin=c]{90}{Fine}}
         & $\text{TL}_{\text{SH}}$~\cite{wu2017sampling} & 66.3 & 33.3 & 34.2 & 32.3 & 35.4 & 48.5 & 54.6 & 68.4 \\
         & NSM~\cite{zhai2018classification} & \underline{70.2} & \textbf{37.6} & \underline{38.0} & 31.4& 28.6 & 36.6 & 43.9 & 63.0 \\
         & NCA++~\cite{teh2020proxynca++} & 67.3 & 37.0 & 37.9 & 33.0 & 32.3 & 41.9 & 48.4 & 66.1 \\
         & \textcolor{black}{Smooth-AP~\cite{smoothap}} & 67.3 & 35.2 & 36.3 & 33.5 & 35.0 & 49.3 & 55.8 & 69.9 \\
         & ROADMAP~\cite{ramzi2021robust} & 69.3 & 35.1 & 35.4 & 29.3 & 29.6 & 46.4 & 54.7 & 69.5 \\
         \midrule
         \multirow{3}{*}{\rotatebox[origin=c]{90}{Hier.}}
         & CSL~\cite{sun2021dynamic} & 59.9 & 30.4 & 32.4 & 36.2 & 50.7 & \underline{81.0} & \underline{87.4} & \underline{91.3} \\
         \cmidrule{2-10}
         & \textbf{\HAPPIER} & \underline{70.2} & 36.0 & 37.0 & \underline{38.0} & \textbf{51.9} & \textbf{81.3} & \textbf{89.1} & \textbf{94.4}  \\
         & \textbf{$\text{\HAPPIER}_{\text{F}}$} & \textbf{70.8} & \textbf{37.6} & \textbf{38.2} & \textbf{38.8} & \underline{50.9} & 76.1 & 82.2 & 83.1 \\
         \bottomrule
    \end{tabularx}
\end{table*}

\begin{table}[t]
\begin{minipage}[t]{0.49\textwidth}
    %
    %
        \caption{Impact of optimization choices for $\hap$ (cf.~\cref{sec:direct_optimization}) on iNat-base.}
     \label{tab:hrank_optim}
    \centering
    \begin{tabularx}{1\textwidth}{YYY }
        \toprule
         $\lhaps$ & \multirow{1}{*}{$\lclust$} & $\hap$ \\
         \midrule
        \xmark & \xmark & 52.3 \\
        \cmark & \xmark & 53.1 \\
        \cmark & \cmark & \textbf{54.3} \\
         \bottomrule
    \end{tabularx}
\end{minipage}%
\hfill
\begin{minipage}[t]{0.49\textwidth}
    %
    %
    \caption{Comparison of $\hap$ (\cref{eq:hierarchy_relevance}) and $\Sigma w\AP$ from supplementary A.2.}
        \label{tab:analysis_relevance}
        \centering
    \begin{tabularx}{1\textwidth}{ l YYY}
        \toprule
         \begin{tabular}{@{}l@{}}
                   test$\rightarrow$\\
                   train$\downarrow$\\
                 \end{tabular} & $\hap$ & $\sum w\AP$ & NDCG \\
         \midrule
         $\hap$ & \textbf{53.1} & 39.8 & \textbf{97.0} \\
        $\sum w\AP$ & 52.0 & \textbf{40.5} & 96.4 \\
         \bottomrule
    \end{tabularx}
\end{minipage}%
\end{table}

\subsection{HAPPIER analysis}\label{sec:happier_analysis}

\medbreak
\noindent\textbf{Ablation study} In~\cref{tab:hrank_optim}, we study the impact of our different choices regarding the direct optimization of $\hap$. The baseline method uses a sigmoid to optimize $\hap$ as in~\cite{Qin2009AGA,smoothap}. Switching to our surrogate loss $\lhaps$ \cref{sec:direct_optimization} yields a +0.8pt increase in $\hap$. Finally, the combination with $\lclust$ in \HAPPIER results in an additional 1.3pt improvement in $\hap$.

\medbreak
\noindent\textbf{Impact of the relevance function} \cref{tab:analysis_relevance} compares models that are trained with the relevance function of~\cref{eq:hierarchy_relevance}, \ie $\hap$, and $\sum w\AP$ (relevance given in supplementary A.2). We report results for $\hap$, $\sum w\AP$ and NDCG. Both $\hap$, $\sum w\AP$ perform better when trained with their own metric: +1.1pt $\hap$ for the model trained to optimize it and +0.7pt $\sum w\AP$ for the model trained to optimize it. Both models show similar performances in NDCG (96.4 \vs 97.0).
\begin{figure}[ht]
    \centering
        
    \begin{subfigure}[t]{0.45\textwidth}
        \includegraphics[width=0.9\textwidth]{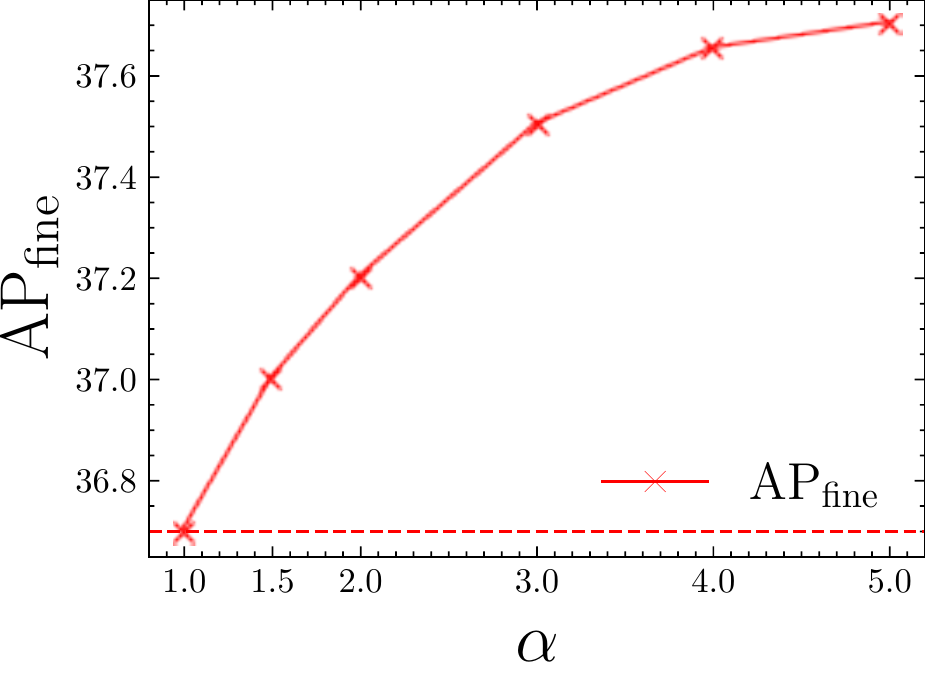}
        \caption{$\AP_{\text{fine}}$ vs $\alpha$ in~\cref{eq:hierarchy_relevance}.
        }
        \label{fig:analysis_alpha}
    \end{subfigure}
    ~
    \begin{subfigure}[t]{0.45\textwidth}
    \includegraphics[width=0.9\textwidth]{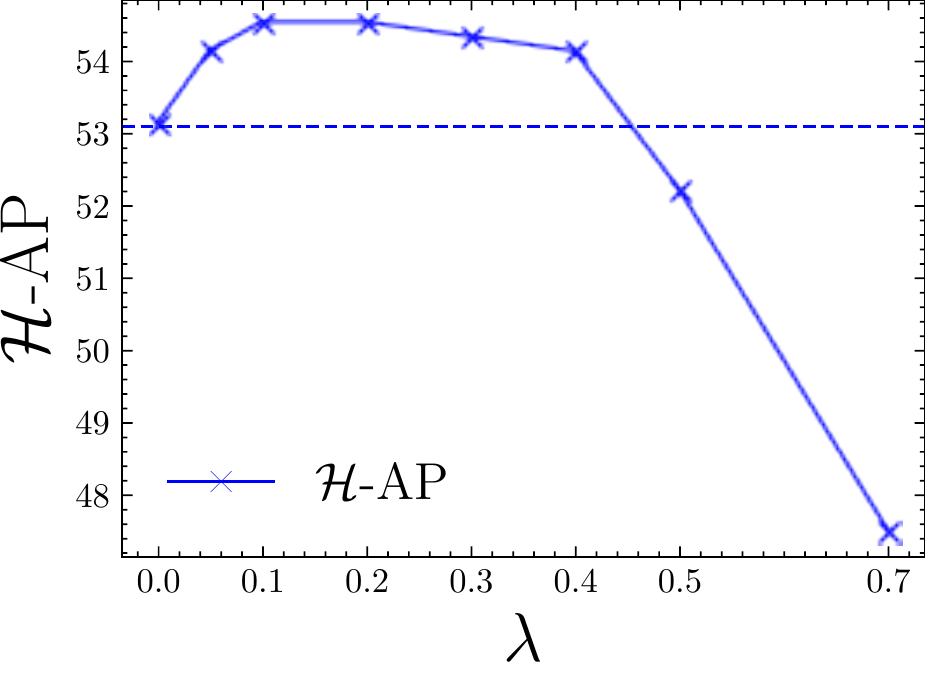}
    \caption{$\hap$ \vs $\lambda$ for $\mathcal{L}_{\text{\HAPPIER}}$.
    }
    \label{fig:analysis_lambda}
    \end{subfigure}%

    \caption{Impact on Inat-base of $\alpha$ in~\cref{eq:hierarchy_relevance} for setting the relevance of $\hap$ (a) and of the $\lambda$ hyper-parameter on \HAPPIER results (b).}
    \label{fig:analysis_alpha_lambda}
\end{figure}

\noindent\textbf{Hyper-parameters} 
\cref{fig:analysis_alpha} studies the impact of $\alpha$ for setting the relevance in~\cref{eq:hierarchy_relevance}: increasing $\alpha$ improves the performances of the AP at the fine-grained level on iNat-base, as expected. We also show in~\cref{fig:analysis_lambda} the impact of $\lambda$ weighting $\lhaps$ and $\lclust$ in \HAPPIER performances: we observe a stable increase in $\hap$ within $0<\lambda<0.5$ compared to optimizing only $\lhaps$, while a drop in performance is observed for $\lambda>0.5$. This shows the complementarity of $\lhaps$ and $\lclust$, and how, when combined, \HAPPIER reaches its best performance.

\subsection{Qualitative study}\label{sec:qualitative_study}

We provide here qualitative assessments of \HAPPIER, including embedding space analysis and visualization of \HAPPIER's retrievals.

\begin{figure}[b]
    \centering
        
    \begin{subfigure}[ht]{0.45\textwidth}
        \includegraphics[width=0.9\textwidth]{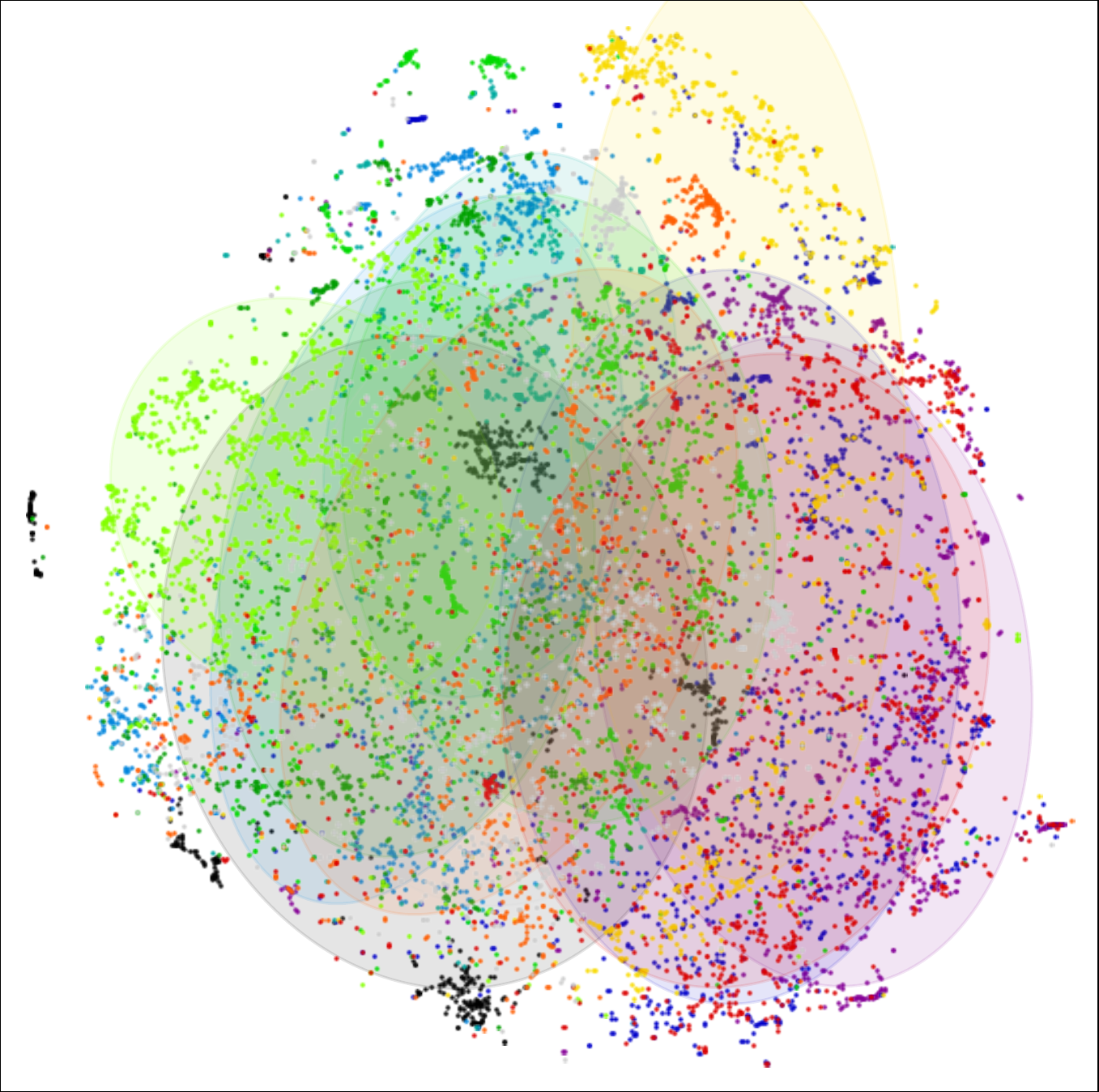}
        \caption{t-SNE visualization of a model trained only on the fine-grained labels.}
        \label{fig:tsne_baseline}
    \end{subfigure}
    ~
    \begin{subfigure}[ht]{0.45\textwidth}
    \includegraphics[width=0.9\textwidth]{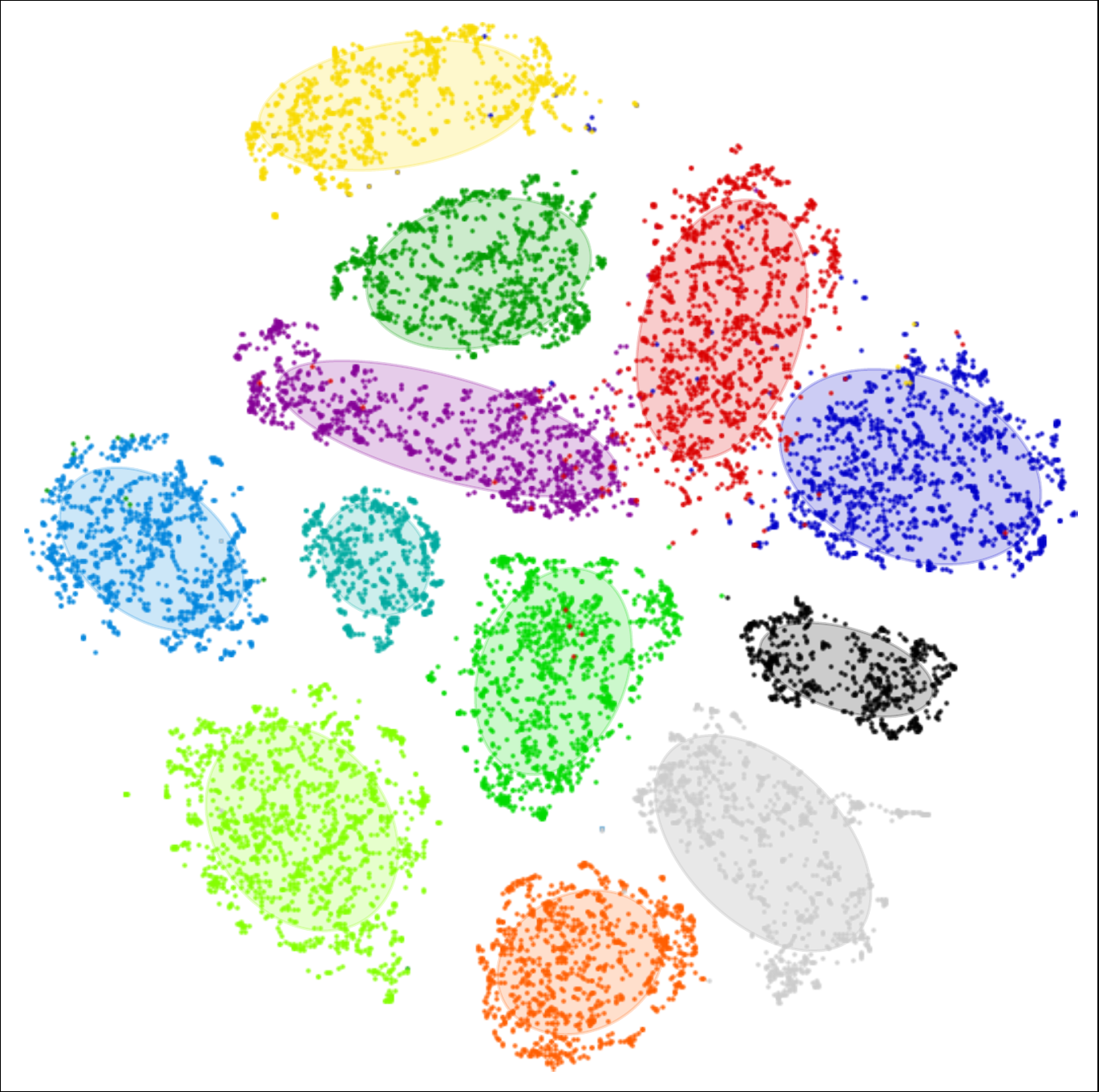}
    \caption{t-SNE visualization of a model trained with \textbf{\HAPPIER}.}
    \label{fig:tsne_happier}
    \end{subfigure}%

    \caption{t-SNE visualisation of the embedding space of two models trained on SOP. Each point is the average embedding of each fine-grained label (object instance) and the colors represent coarse labels (object category, \eg bike, coffee maker).}
    \label{fig:tsne}
\end{figure}

\medbreak
\noindent\textbf{t-SNE: organization of the embedding space} In~\cref{fig:tsne}, we plot using t-SNE~\cite{tsne,chan2019gpu} how \HAPPIER learns an embedding space on SOP ($L=2$) that is well-organized. We plot the mean vector of each fine-grained class and we assign the color based on the coarse level. We show on~\cref{fig:tsne_baseline} the t-SNE visualisation obtained using a baseline method trained on the fine-grained labels, and in~\cref{fig:tsne_happier} we plot the t-SNE of the embedding space of a model trained with \HAPPIER. We cannot observe any clear clusters for the coarse level on~\cref{fig:tsne_baseline}, whereas we can appreciate the the quality of the hierarchical clusters formed on~\cref{fig:tsne_happier}.

\begin{figure}[t]
    \centering
        
    \begin{subfigure}[t]{\textwidth}
        \includegraphics[width=0.9\textwidth]{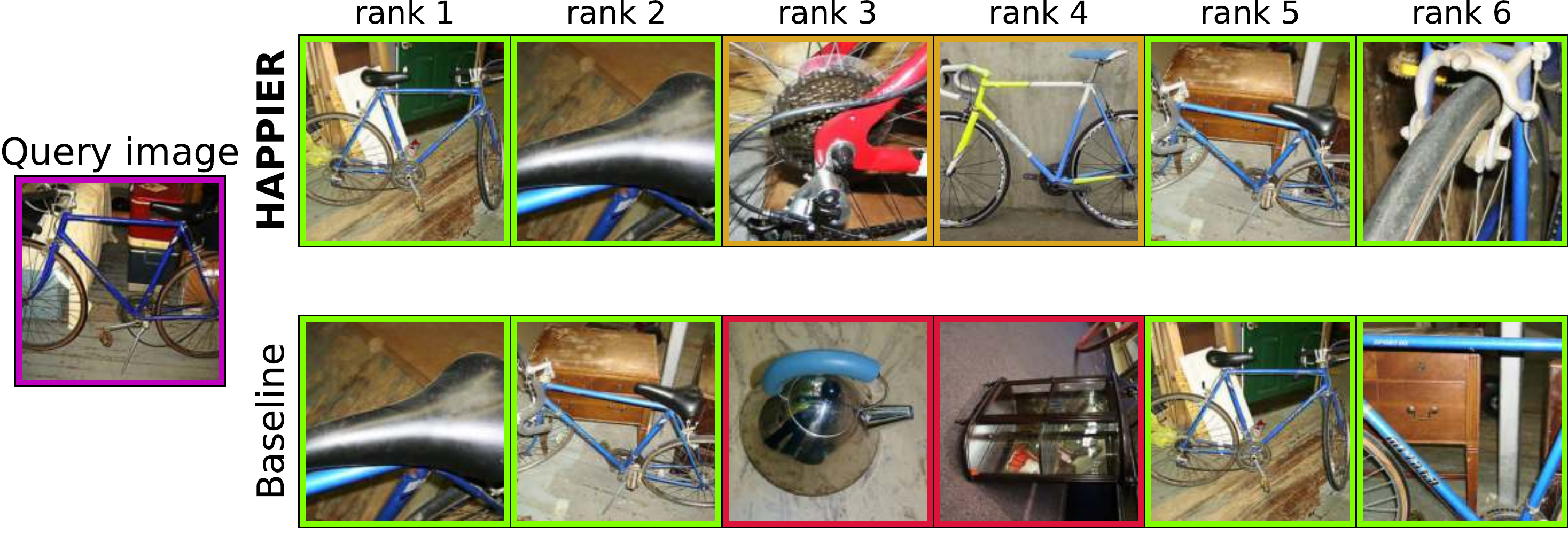}
        \caption{\HAPPIER can help make less severe mistakes. The inversion on the bottom row are with negative instances (in \textcolor{red}{red}), where as with \HAPPIER (top row) inversions are with instances sharing the same coarse label ``bike'' (in \textcolor{orange}{orange}).}
        \label{fig:qual_sop_good}
    \end{subfigure}
    
    \begin{subfigure}[t]{\textwidth}
    \includegraphics[width=0.9\textwidth]{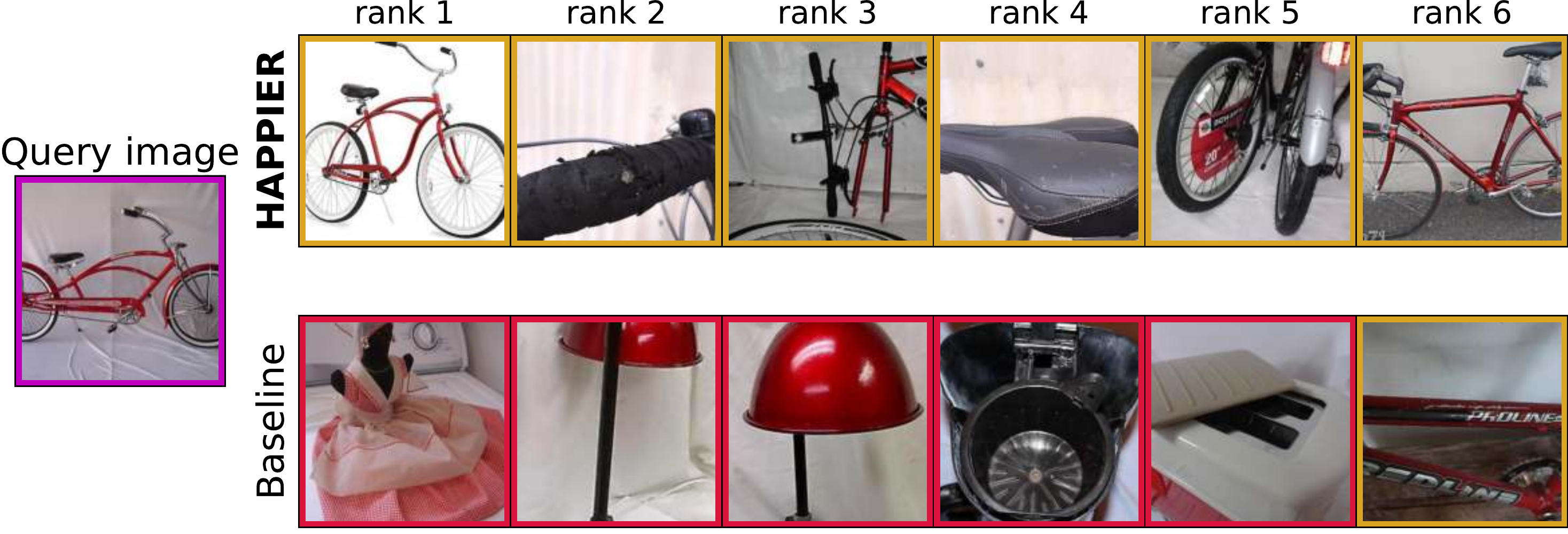}
    \caption{In this example, the models fail to retrieve the correct fine grained images. However \HAPPIER still retrieves images of very similar bikes (in \textcolor{orange}{orange}) whereas the baseline retrieves images that are dissimilar semantically to the query (in \textcolor{red}{red}).}
    \label{fig:qual_sop_error}
    \end{subfigure}%

    \caption{Qualitative examples of failure cases from a standard fine-grained model corrected by training with \HAPPIER.}
    \label{fig:qualitative_results}
\end{figure}

\medbreak
\noindent\textbf{Controlled errors} Finally, we showcase in~\cref{fig:qualitative_results} errors of \HAPPIER \vs a fine-grained baseline. On~\cref{fig:qual_sop_good}, we illustrate how a model trained with \HAPPIER makes mistakes that are less severe than a baseline model trained only on the fine-grained level. On~\cref{fig:qual_sop_error}, we show an example where both models fail to retrieve the correct fine-grained instances, however the model trained with \HAPPIER retrieves images of bikes that are visually more similar to the query.

%% file: files/conclusion.tex
\section{Conclusion}\label{sec:conclusion}

In this work, we introduce \HAPPIER, a new training method that leverages hierarchical relations between concepts to learn robust rankings. \HAPPIER is based on a new metric $\hap$ that evaluates hierarchical rankings and uses a combination of a smooth upper bound surrogate with theoretical guarantees and a clustering loss to directly optimize it. Extensive experiments show that \HAPPIER performs on par to state-of-the-art image retrieval methods on fine-grained metrics and exhibits large improvements \vs recent hierarchical methods on hierarchical metrics. Learning more robust rankings reduces the severity of ranking errors, and is qualitatively related to a better organization of the embedding space with \HAPPIER. Future works include the adaptation of \HAPPIER to the unsupervised setting, \eg for providing a relevant self-training criterion.

\medbreak
\noindent\textbf{Acknowledgement} This work was done under a grant from the the AHEAD ANR program (ANR-20-THIA-0002). It was granted access to the HPC resources of IDRIS under the allocation 2021-AD011012645 made by GENCI.

%% file: supplementary_file.tex
\setcounter{section}{0}
\renewcommand\thesection{\Alph{section}}

\section{Method}

\subsection{$\hrank$}\label{sec:sup_hrank}

We define the $\hrank$ in the main paper as:

\begin{equation}
    \hrank(k) = \rel(k) + \sum_{j\in\Omega^+} \min(\rel(k), \rel(j))\cdot H(s_j-s_k) ~.
    \label{eq:sup_hierarchical_rank}
\end{equation}

We detail in~\cref{fig:supp_hrank_figure} how the $\hrank$ in~\cref{eq:sup_hierarchical_rank} is computed in the example from Fig.~2b of the main paper. Given a ``Lada \#2'' query, we set the relevances as follows: if $k\in\Omega^{(3)}$ (\ie $k$ is also a ``Lada \#2''), $\rel(k)=1$; if $k\in\Omega^{(2)}$ (\ie $k$ is another model of ``Lada''), $\rel(k)=2/3$; and if $k\in\Omega^{(1)}$ ($k$ is a ``Car''), $\rel(k)=1/3$. Relevance of negatives (other vehicles) is set to 0.

\definecolor{amethyst}{rgb}{0.6, 0.4, 0.8}
\begin{figure}
    \centering
    \includegraphics[width=\textwidth]{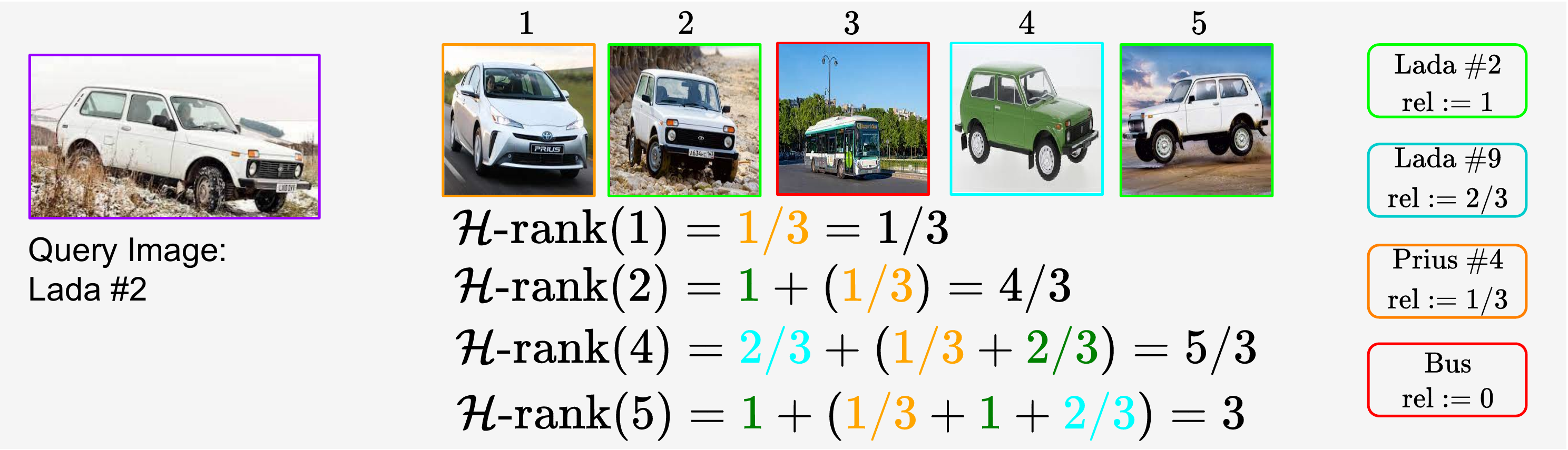}
    \caption{$\hrank$ for each retrieval results given a ``\textcolor{Green}{Lada \#2}'' \textcolor{amethyst}{query} with relevances of~\cref{sec:sup_hrank} and the hierarchical tree of Fig.~2a of the main paper.}
    \label{fig:supp_hrank_figure}
\end{figure}

In this instance, $\hrank(2)=4/3$ because $\rel(2)=1$ and $\min(\rel(1), \rel(2)) = \rel(1) = 1/3$. Here, the closest common ancestor in the hierarchical tree shared by the query and instances $1$ and $2$ is ``Cars''. For binary labels, we would have $\rank^+(2)=1$; this would not take into account the semantic similarity between the query and instance $1$.

\subsection{$\hap$}

We define $\hap$ in the main paper as:

\begin{equation}\label{eq:sup_def_hap}
    \hap = \frac{1}{\sum_{k\in\Omega^+}\rel(k)} \sum_{k\in\Omega^+} \frac{\hrank(k)}{\rank(k)}
\end{equation}

\begin{figure}[t]
    \centering
    \includegraphics[width=\textwidth]{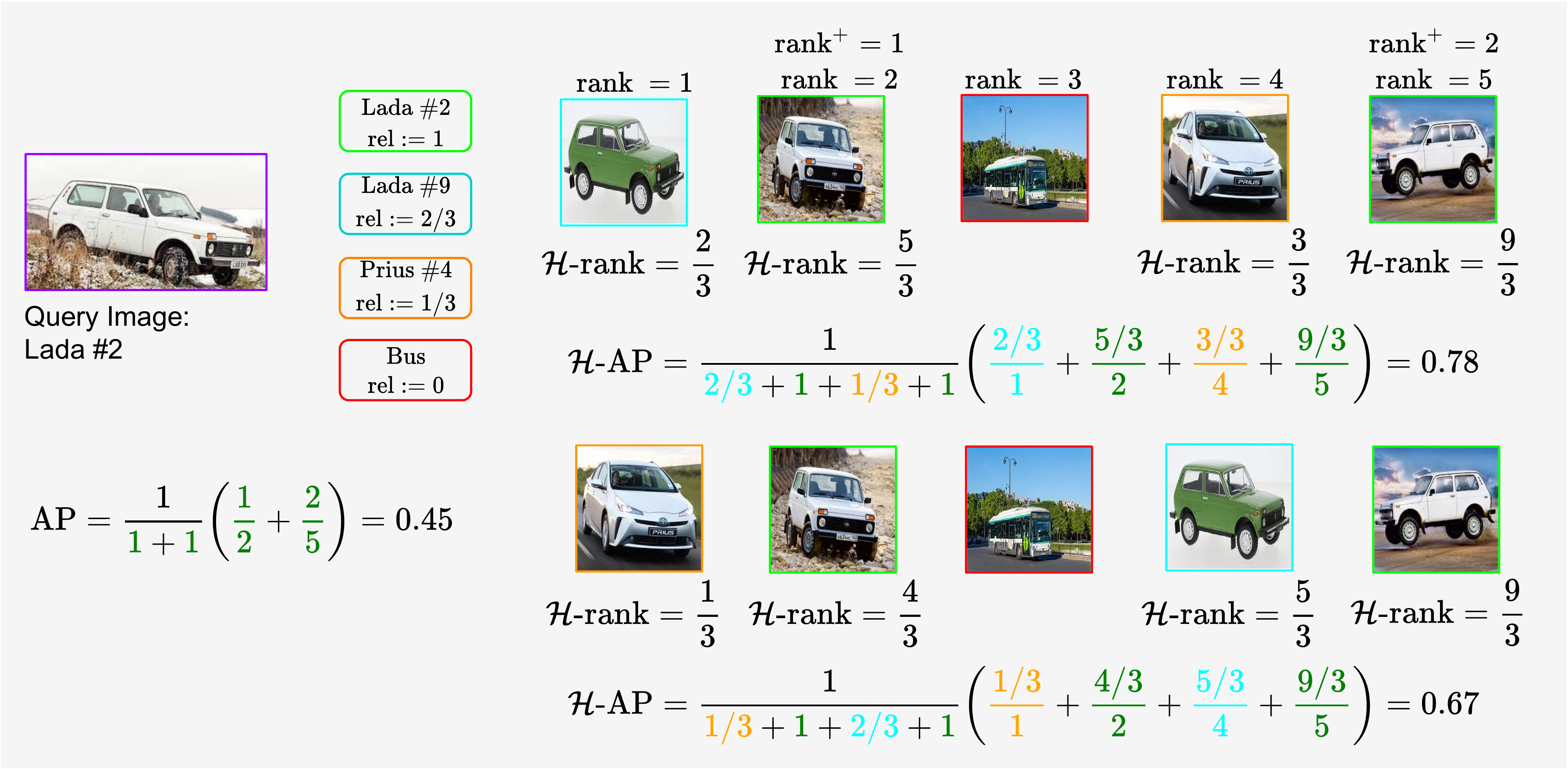}
    \caption{AP and $\hap$ for two different rankings when Given a ``\textcolor{Green}{Lada \#2}'' \textcolor{amethyst}{query} and relevances of~\cref{sec:sup_hrank}. The $\hap$ of the top row is greater (0.78) than the bottom one's (0.67) as the error in $\rank=1$ is less severe for the top row. Whereas the AP is the same for both rankings (0.45).}
    \label{fig:sup_dif_ap_hap}
\end{figure}

We illustrate in~\cref{fig:sup_dif_ap_hap} how the $\hap$ is computed for both rankings of Fig.~2b of the main paper. We use the same relevances as in~\cref{sec:sup_hrank}. The $\hap$ of the first example is greater ($0.78$) than of the second one ($0.67$) because the error is less severe. On the contrary, the AP only considers binary labels and is the same for both rankings ($0.45$).

\medbreak

\textcolor{black}{One property of AP is that it can be interpreted as the area under the precision-recall curve. $\hap$ from~\cref{eq:sup_def_hap} can also be interpreted as the area under a hierarchical-precision-recall curve by defining a Hierarchical Recall ($\mathcal{H}\text{-R@k}$) and a Hierarchical Precision ($\mathcal{H}\text{-P@k}$) as:}

\begin{align}
    &\mathcal{H}\text{-R@k} = \frac{\sum_{j=1}^k \rel(j)}{\sum_{j\in\Omega^+} \rel(j)} \label{eq:hierarchical_recall}\\
    &\mathcal{H}\text{-P@k} = \frac{\sum_{j=1}^k \min(\rel(j),\rel(k))}{k\cdot\rel(k)} \label{eq:hierarchical_precision}
\end{align}

\textcolor{black}{So that $\hap$ can be re-written as:}
\begin{equation}
\label{eq:hap_generalizes_ap}
    \hap = \sum_{k=1}^{|\Omega|} (\mathcal{H}\text{-R@k}-\mathcal{H}\text{-R@k-1}) \times \mathcal{H}\text{-P@k}
\end{equation}

\textcolor{black}{\cref{eq:hap_generalizes_ap} recovers Eq. (3) from the main paper, meaning that $\hap$ generalizes this property of AP beyond binary labels. To further motivate $\hap$ we will justify the normalization constant for $\hap$, and show that $\hap$, $\mathcal{H}\text{-R@k}$ and $\mathcal{H}\text{-P@k}$ are consistent generalization of AP, R@k, P@k.}

\subsubsection{Normalization constant for $\hap$} When all instances are perfectly ranked, all instances $j$ that are ranked before instance $k$ ($s_j\geq s_k$) have a relevance that is higher or equal than $k$'s, \ie $\rel(j)\geq\rel(k)$ and \\ ${\min(\rel(j),\rel(k))=\rel(k)}$. So, for each instance $k$:

\begin{align*}
    \hrank(k) &= \rel(k) + \sum_{j\in\Omega^+} \min(\rel(k), \rel(j))\cdot H(s_j-s_k) \\
    &= \rel(k) + \sum_{j\in\Omega^+} \rel(k)\cdot H(s_j-s_k) \\
    &= \rel(k) \cdot \left( 1 + \sum_{j\in\Omega^+} H(s_j-s_k) \right) = \rel(k)\cdot\rank(k)
\end{align*}
The total sum $\sum_{k\in\Omega^+} \frac{\hrank(k)}{\rank(k)} = \sum_{k\in\Omega^+} \rel(k)$. This means that we need to normalize by $\sum_{k\in\Omega^+} \rel(k)$ in order to constrain $\hap$ between 0 and 1. This results in the definition of $\hap$ from~\cref{eq:sup_def_hap}.

\subsubsection{$\hap$ is a consistent generalization of AP} In a binary setting, AP is defined as follows:

\begin{equation}
    \AP = \frac{1}{|\Omega^+|} \sum_{k\in\Omega^+} \frac{\rank^+(k)}{\rank(k)}
\end{equation}

$\hap$ is equivalent to AP in a binary setting ($L=1$). Indeed, the relevance function is $1$ for fine-grained instances and 0 otherwise in the binary case. Therefore $\hrank(k) = 1 + \sum_{j\in\Omega^+} H(s_j-s_k)$ which is the same definition as $\rank^+$ in AP. Furthermore the normalization constant of $\hap$, $\sum_{k\in\Omega^+} \rel(k)$, is equal to the number of fine-grained instances in the binary setting, \ie $|\Omega^+|$. This means that $\hap=\AP$ in this case.

\medbreak
\textcolor{black}{$\mathcal{H}\text{-R@k}$ is also a consistent generalization of R@k, indeed: $$\mathcal{H}\text{-R@k} = \frac{\sum_{j=1}^k \rel(j)}{\sum_{j\in\Omega^+} \rel(j)} = \frac{\sum_{j=1}^k \mathds{1}(k\in\Omega^+)}{\sum_{j\in\Omega^+} \mathds{1}(k\in\Omega^+)} = \frac{\text{\# number of positive before k}}{|\Omega^+|} = R@k$$}

\medbreak
\textcolor{black}{Finally, $\mathcal{H}\text{-P@k}$ is also a consistent generalization of P@k:\\ $$\mathcal{H}\text{-P@k} = \frac{\sum_{j=1}^k \min(\rel(j),\rel(k))}{k\cdot\rel(k)}=\frac{\text{\# number of positive before k}}{k}=P@k$$}

\subsubsection{Link between $\hap$ and the weighted average of AP}

Let us define the AP for the semantic level $l\geq1$ as the binary AP with the set of positives being all instances that belong the same level, \ie $\Omega^{+,l} = \bigcup_{q=l}^L \Omega^{(q)}$:

\begin{equation}\label{eq:sup_ap_level}
    \AP^{(l)} = \frac{1}{|\Omega^{+,l}|} \sum_{k\in\Omega^{+,l}} \frac{\rank^{+,l}(k)}{\rank(k)}, \; \rank^{+,l}(k) = 1 + \sum_{j\in\Omega^{+,l}} H(s_j-s_k)
\end{equation}

\fbox{
\parbox{0.9\linewidth}{%
\begin{property}\label{prop:link_hap_ap} For any relevance function  $\rel(k) = \sum_{p=1}^l \frac{w_p}{|\Omega^{+,q}|}, \, k\in\Omega^{(l)}$, with positive weights $\{w_l\}_{l \in \llbracket 1;L\rrbracket}$ such that $\sum_{l=1}^L w_l =1$:

\begin{equation*}
    \hap = \sum_{l=1}^L w_l \cdot AP^{(l)}
\end{equation*}

\ie $\hap$ is equal the weighted average of the AP at all semantic levels.
\end{property}
}
}

\medbreak
\medbreak
\textbf{Proof of Property~\ref{prop:link_hap_ap}}

\medbreak
Denoting $\Sigma w\AP:=\sum_{l=1}^L w_l \cdot AP^{(l)}$, we obtain from~\cref{eq:sup_ap_level}:

\begin{equation}
    \Sigma w\AP = \sum_{l=1}^L  w_l \cdot \frac{1}{|\Omega^{+,l}|} \sum_{k\in\Omega^{+,l}} \frac{\rank^{+,l}(k)}{\rank(k)}
\end{equation}

We define $\hat{w}_l = \frac{w_l}{|\Omega^{+,l}|}$ to ease notations, so:

\begin{equation}
    \Sigma w\AP = \sum_{l=1}^L  \hat{w}_l \sum_{k\in\Omega^{+,l}} \frac{\rank^{+,l}(k)}{\rank(k)} 
\end{equation}

We define $\mathds{1}(k,l) = \mathds{1}\left[k\in\Omega^{+,l} \right]$ so that we can sum over $\Omega^+$ instead of $\Omega^{+,l}$ and inverse the summations. Note that rank does not depend on $l$, on contrary to $\rank^{+,l}$.
\begin{align}
    \Sigma w\AP &= \sum_{l=1}^L \sum_{k\in\Omega^+} \frac{\hat{w}_l \cdot \mathds{1}(k,l) \cdot \rank^{+,l}(k)}{\rank(k)} \\
    &= \sum_{k\in\Omega^+} \sum_{l=1}^L \frac{\hat{w}_l \cdot \mathds{1}(k,l) \cdot \rank^{+,l}(k)}{\rank(k)} \\
    &= \sum_{k\in\Omega^+} \frac{\sum_{l=1}^L  \mathds{1}(k,l) \cdot \hat{w}_l \cdot \rank^{+,l}(k)}{\rank(k)} \label{eq:it_is_inverted}
\end{align}

We replace $\rank^{+,l}$ in~\cref{eq:it_is_inverted} with its definition from~\cref{eq:sup_ap_level}:
\begin{align}
    \Sigma w\AP &= \sum_{k\in\Omega^+} \frac{\sum_{l=1}^L  \mathds{1}(k,l) \cdot \hat{w}_l \cdot \left(1 + \sum_{j\in\Omega^{+,l}} H(s_j-s_k)\right)}{\rank(k)} \\
    &= \sum_{k\in\Omega^+} \frac{\sum_{l=1}^L  \mathds{1}(k,l) \cdot \hat{w}_l + \sum_{l=1}^L \sum_{j\in\Omega^{+,l}} \mathds{1}(k,l) \cdot \hat{w}_l \cdot H(s_j-s_k)}{\rank(k)} \\
    &= \sum_{k\in\Omega^+} \frac{\sum_{l=1}^L  \mathds{1}(k,l) \cdot \hat{w}_l +  \sum_{l=1}^L \sum_{j\in\Omega^+} \mathds{1}(j,l) \cdot \mathds{1}(k,l) \cdot \hat{w}_l \cdot H(s_j-s_k)}{\rank(k)} \\
    &= \sum_{k\in\Omega^+} \frac{\sum_{l=1}^L  \mathds{1}(k,l) \cdot \hat{w}_l + \sum_{j\in\Omega^+} \sum_{l=1}^L \mathds{1}(j,l) \cdot \mathds{1}(k,l) \cdot \hat{w}_l \cdot H(s_j-s_k)}{\rank(k)} \label{eq:nearl_hrank}
\end{align}

We define the following relevance function:
\begin{equation}\label{eq:temporary_relevance}
    \rel(k)=\sum_{l=1}^L  \mathds{1}(k,l) \cdot \hat{w}_l
\end{equation}
By construction of $\mathds{1}(\cdot,l)$: 
\begin{equation}\label{eq:see_the_min}
    \sum_{l=1}^L \mathds{1}(j,l) \cdot \mathds{1}(k,l) \cdot \hat{w}_l = \min(\rel(k), \rel(j))
\end{equation}

Using the definition of the relevance function from~\cref{eq:temporary_relevance} and~\cref{eq:see_the_min}, we can rewrite~\cref{eq:nearl_hrank} with $\hrank$:

\begin{align}
    \Sigma w\AP &= \sum_{k\in\Omega^+} \frac{\rel(k) + \sum_{j\in\Omega^+} \min(\rel(j), \rel(k)) \cdot H(s_j-s_k)}{\rank(k)} \\
    &= \sum_{k\in\Omega^+} \frac{\hrank(k)}{\rank(k)} \label{eq:almost_hap}
\end{align}

\cref{eq:almost_hap} lacks the normalization constant $\sum_{k\in\Omega^+} \rel(k)$ in order to have the same shape as $\hap$ in~\cref{eq:sup_def_hap}. So we must prove that $\sum_{k\in\Omega^+} \rel(k) = 1$:
\begin{align}
    \sum_{k\in\Omega^+} \rel(k) &= \sum_{k\in\Omega^+} \sum_{l=1}^L  \mathds{1}(k,l) \cdot \hat{w}_l \\
    &= \sum_{l=1}^L |\Omega^{(l)}| \sum_{p=1}^l \hat{w}_p \\
    &= \sum_{l=1}^L |\Omega^{(l)}| \sum_{p=1}^l \frac{w_p}{|\Omega^{+,p}|} \\
    &= \sum_{l=1}^L |\Omega^{(l)}| \sum_{p=1}^l \frac{w_p}{|\bigcup_{q=p}^L \Omega^{(q)}|} \\
    &= \sum_{l=1}^L |\Omega^{(l)}| \sum_{p=1}^l \frac{w_p}{\sum_{q=p}^L |\Omega^{(q)}|} \\
    &= \sum_{l=1}^L  \sum_{p=1}^l \frac{|\Omega^{(l)}| \cdot w_p}{\sum_{q=p}^L |\Omega^{(q)}|} \\
    &= \sum_{p=1}^L  \sum_{l=p}^L \frac{|\Omega^{(l)}| \cdot w_p}{\sum_{q=p}^L |\Omega^{(q)}|} \\
    &= \sum_{p=1}^L w_p \cdot \frac{ \sum_{l=p}^L |\Omega^{(l)}| }{\sum_{q=p}^L |\Omega^{(q)}|} \\
    &= \sum_{p=1}^L w_p = 1
\end{align}

We have proved that $\Sigma w\AP = \hap$ with the relevance function of~\cref{eq:temporary_relevance}:
\begin{equation}
    \Sigma w\AP = \frac{1}{\sum_{k\in\Omega^+}\rel(k)} \sum_{k\in\Omega^+} \frac{\hrank(k)}{\rank(k)} = \hap
\end{equation}

Finally we show, for an instance $k\in\Omega^{(l)}$, :

\begin{equation}
    \rel(k)=\sum_{p=1}^L  \mathds{1}(k,p) \cdot \hat{w}_p = \sum_{p=1}^l \cdot \hat{w}_p = \sum_{p=1}^l \frac{w_p}{|\Omega^{+,p}|}
\end{equation}
\ie the relevance of~\cref{eq:temporary_relevance} is the same as the relevance of Property~\ref{prop:link_hap_ap}. This concludes the proof of Property~\ref{prop:link_hap_ap}. $\square$

\subsection{Direct optimisation of $\hap$}

\subsubsection{Decomposing $\hrank$ and $\rank$}

We have $\Omega^+ = \bigcup_{q=1}^L \Omega^{(q)}$, for an instance $k\in\Omega^{(l)}$ we can define the following subsets: $\Omega^> = \bigcup_{q=l+1}^L \Omega^{(q)}$ and $\Omega^\leq = \bigcup_{q=1}^l \Omega^{(q)}$, so that $\Omega^+=\Omega^>\cup\Omega^\leq$. So we can rewrite $\hrank$:

\begin{align*}
    \hrank(k) &= \rel(k) + \sum_{j\in\Omega^+} \min(\rel(k), \rel(j))\cdot H(s_j-s_k) \\
    &= \underbrace{\sum_{j\in\Omega^>} \min(\rel(k), \rel(j))\cdot H(s_j-s_k)}_{\hrank^>} \\&+ \underbrace{\rel(k) + \sum_{j\in\Omega^\leq} \min(\rel(k), \rel(j))\cdot H(s_j-s_k)}_{\hrank^\leq}
\end{align*}

Similarly we can define $\Omega^\geq = \bigcup_{q=l}^L \Omega^{(q)}$ and $\Omega^< = \bigcup_{q=0}^{l-1} \Omega^{(q)}$, with $\Omega^+=\Omega^\geq\cup\Omega^<$. So we can rewrite $\rank$:

\begin{align*}
    \rank(k) &= 1 + \sum_{k\in\Omega} H(s_j-s_k) \\
    &= \underbrace{1 + \sum_{k\in\Omega^\geq} H(s_j-s_k)}_{\rank^\geq} + \underbrace{\sum_{k\in\Omega^<} H(s_j-s_k)}_{\rank^<}
\end{align*}

\subsubsection{Gradients for $\lhap$} We further decompose $\lhap$ from Eq.~5 of the main paper, using $\hrank^\leq(k) = \hrank^=(k) + \hrank^<(k)$, $\rank^\geq(k) = \rank^>(k) + \rank^=(k)$:

\begin{equation*}\label{eq:rewrite_hap_sup}
    \lhap = 1 - \frac{1}{\sum_{k\in\Omega^+}\rel(k)} \sum_{k\in\Omega^+} \frac{\hrank^>(k) + \hrank^=(k) + \hrank^<(k)}{\rank^>(k) + \rank^=(k) + \rank^<(k) + \rank^-(k)}
\end{equation*}

\begin{table}[ht]
    \caption{Decomposition of $\hap$ for optimization.}
    \label{tab:sup_choice_optim}
    \centering
    \begin{tabularx}{1\textwidth}{l YYYYYYY }
        \toprule
          & $\hrank^>$ & $\rank^<$ & $\rank^-$ & $\hrank^=$ & $\hrank^<$ & $\rank^>$ & $\rank^=$ \\
         \midrule
        Optimization & \textcolor{blue}{\cmark} & \textcolor{blue}{\cmark} & \textcolor{blue}{\cmark} & \textcolor{red}{\xmark} & \textcolor{red}{\xmark} & \textcolor{red}{\xmark} & \textcolor{red}{\xmark}  \\
         \bottomrule
    \end{tabularx}
\end{table}

We choose to only optimize with respect to the terms indicated with \textcolor{blue}{\cmark} in~\cref{tab:sup_choice_optim}.

\medbreak
\textbf{$\boldsymbol{\rank^-(k)}$}: $\frac{\partial \lhap}{\partial \rank^-(k)} \propto \frac{\hrank(k)}{\rank(k)^2} > 0$ which means that in order to decrease $\lhap$ we must lower $\rank^-$, which is an expected behaviour, as it will force $k$ to have a better ranking if it ranked after negative instances (in $\Omega^-$).

\medbreak
\textbf{$\boldsymbol{\rank^<(k)}$}: if we suppose that $\hrank^<$ is a constant, then $\frac{\partial \lhap}{\partial \rank^<(k)} \propto \frac{\hrank(k)}{\rank(k)^2} > 0$ which means that in order to decrease $\lhap$ we must lower $\rank^<$, which is an expected behaviour, as it will force $k$ to have a better ranking if it ranked after negative instances (in $\Omega^<$).

\medbreak
\textbf{$\boldsymbol{\hrank^>(k)}$}: if we suppose that $\rank^>$ is a constant, $ \frac{\partial \lhap}{\partial \hrank^>(k)} \propto \frac{-1}{\rank(k)} < 0$ which means that in order to decrease $\lhap$ we must increase $\hrank^>$, which is an expected behaviour, as it will force $k$ to be ranked after other instances of higher relevance (in $\Omega^>$).

\medbreak
We choose to not optimize with respect to $\hrank^=$, $\hrank^<$, $\rank^>$, $\rank^=$.

\medbreak

\textbf{$\boldsymbol{\rank^=}$ \& $\boldsymbol{\hrank^=}$}: Optimizing through $\rank^=$ has no impact so we choose not to optimize it, indeed $\frac{\partial \lhap}{\partial \rank^=} = 0$. This is the case because inversions between instances of same relevance has no impact on $\hap$. This is also the case for $\hrank^=$.

\medbreak

\textbf{$\boldsymbol{\hrank^<(k)}$}: $\hrank^<(k)$ depends on $\rank^<(k)$ and the relevance of the other instances that are before. We note that $0<\frac{\partial \hrank^<(k)}{\partial \rank(k)} < \rel(k)$ indeed when the $\rank^<$ increases $\hrank^<$ increases and the increase rate can not be equal or greater than $\rel(k)$

\begin{align}
    \frac{\partial \lhap}{\partial \rank^<(k)} \propto & -\Bigg( \overbrace{\left(\frac{\partial \hrank^<(k)}{\partial \rank^<(k)}-\rel(k)\right)\cdot\rank^>(k)}^{a} \\
    &+ \overbrace{\left(\frac{\partial \hrank^<(k)}{\partial \rank^<(k)}-\rel(k)\right)\cdot\rank^=(k)}^{b} \\
    &+ \overbrace{\left(\frac{\partial \hrank^<(k)}{\partial \rank^<(k)}\cdot\rank^<(k)-\hrank^<(k)\right)}^{c} \\
    &+ \overbrace{\frac{\partial \hrank^<(k)}{\partial \rank^<(k)}\cdot\rank^-(k)}^{d}\Bigg)/\rank(k)^2
\end{align}

When optimizing through $\hrank^<$ we can no longer explicitly control the sign of $\frac{\partial \lhap}{\partial \rank^<(k)}$. For example if $a$ and $b$ are null (\ie not instances of higher or equal relevance are above $k$), $d$ remains and is greater than $0$ and $c$ can be greater than $0$ resulting in an overall negative gradient, which is an unexpected behaviour. This is why we choose to not optimize through $\hrank^<$.

\medbreak

\textbf{$\boldsymbol{\rank^>(k)}$}: We have $\hrank^>(k)=\rel(k)\cdot\rank^>(k)$ indeed all instances $j$ ranked before $k$ have a strictly higher relevance, \ie $\min(\rel(j), \rel(k))=\rel(k)$, so we can write:

\begin{equation}
    \frac{\partial \lhap}{\partial \rank^>(k)} \propto \frac{\overbrace{\hrank^<(k) - \rel(k)\cdot\rank^<(k)}^{<0} -\rel(k)\cdot\rank^-(k) }{\rank(k)^2} < 0
\end{equation}

Optimizing trough $\rank^>$ instead of only $\hrank^>$ diminishes the magnitude of the resulting gradient, so we decide to not optimize through $\rank^>$.

\subsubsection{Approximating $\hrank^>$} In order to have a lower bound on $\hrank^>$ we approximate the Heaviside step function $H$ with a smooth lower bound:

\begin{equation}
    H^>_s(t) = 
    \begin{cases}
      \gamma \cdot t, \quad \text{if } t < 0 \\
      \max(\nu\cdot t + \mu, 1), \quad \text{if } t \geq 0 \\
    \end{cases}
    \label{eq:sup_h_uparrow}
\end{equation}
$H^>_s$ is illustrated in~\cref{fig:sup_hrank_surrogate}. Using $H^>_s$ we can approximate $\hrank^>$: $\hrank^>_s(k) = \rel(k) + \sum_{j\in\Omega^+} \min(\rel(j), \rel(k)) H^>_s(s_j-s_k)$. Because $H^>_s(t) \leq H(t)$: $\hrank^>_s(k) \leq \hrank^>$. In our experiments we use: $\gamma=10$, $\nu=25$, $\mu=0.5$.

\subsubsection{Approximating $\rank^<$} 
\begin{figure}[ht!]
    \centering
        
    \begin{subfigure}[t]{0.45\textwidth}
        \includegraphics[width=1\textwidth]{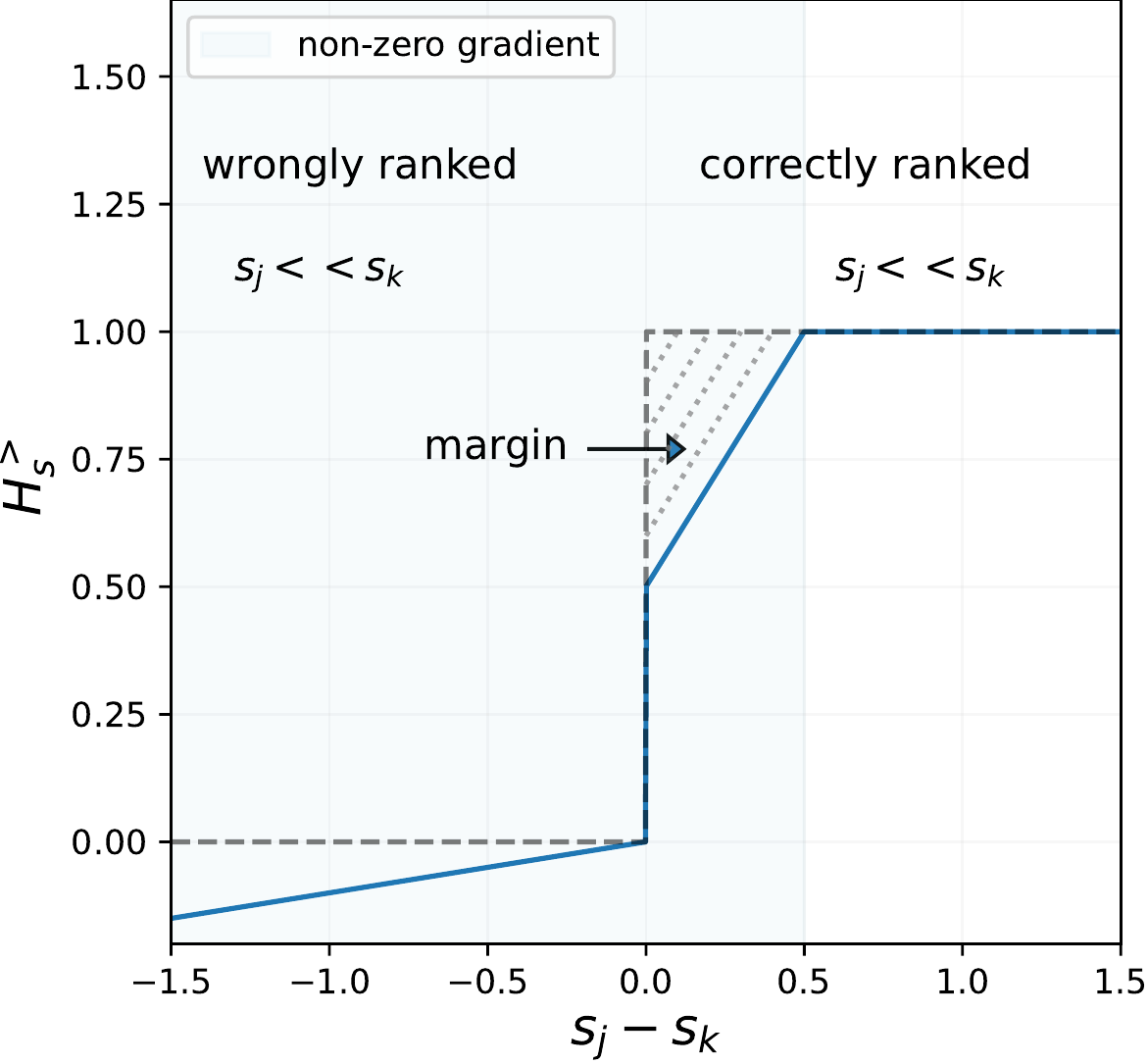}
        \caption{Illustration of $H^>_s$ in~\cref{eq:sup_h_uparrow}.}
        \label{fig:sup_hrank_surrogate}
    \end{subfigure}
    ~
    \begin{subfigure}[t]{0.45\textwidth}
    \includegraphics[width=1\textwidth]{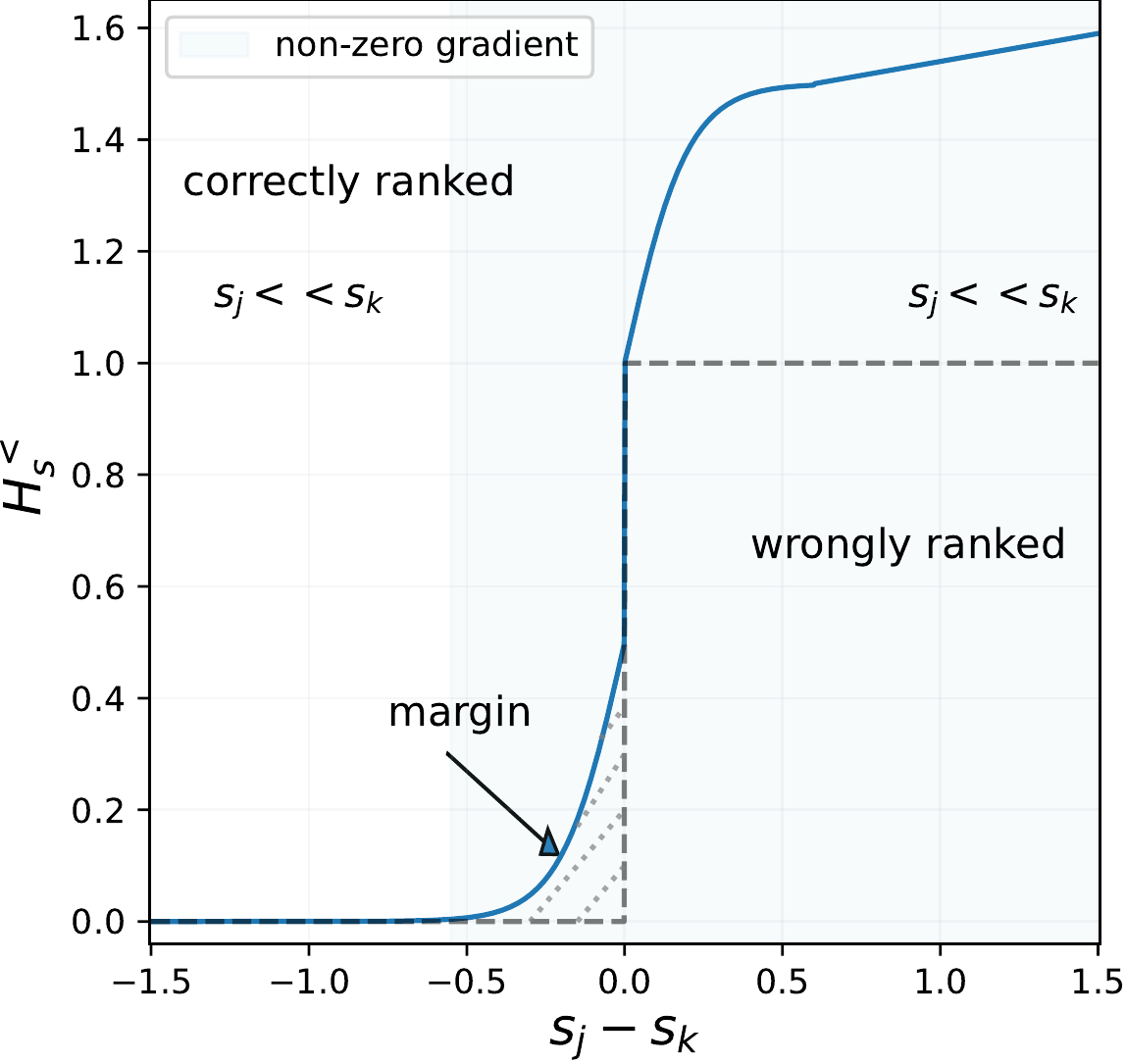}
    \caption{Illustration of $H^<_s$ in~\cref{eq:sup_h_uparrow}.}
    \label{fig:sup_rank_surrogate}
    \end{subfigure}%

    \caption{Illustrations of the two approximations of the Heaviside step function used to approximate $\hrank^>$ and $\rank^<$.}
    \label{fig:sup_surrogates}
\end{figure}

In order to have an upper bound on $\rank^<$ we approximate the Heaviside with a smooth upper bound as given in~\cite{ramzi2021robust}:

\begin{equation}
    H^<_s(t) = 
    \begin{cases}
      \sigma(\frac{t}{\tau}) \quad \text{if} \; t \leq 0, \quad \text{where $\sigma$ is the sigmoid function} \\
      \sigma(\frac{t}{\tau}) + 0.5 \quad \text{if} \; t \in [0;\delta] \quad \text{with} \; \delta \geq 0\\
      \rho \cdot (t - \delta) + \sigma(\frac{\delta}{\tau}) + 0.5 \quad \text{if} \; t > \delta \\
    \end{cases}
    \label{eq:sup_h_minus}
\end{equation}
$H^<_s$ is illustrated in~\cref{fig:sup_hrank_surrogate}. Using $H^<_s$ we can approximate $\rank^<$: $\rank^<_s(k) = 1 + \sum_{j\in\Omega} H^<_s(s_j-s_k)$. Because $H^<_s(t) \geq H(t)$: $\rank^<_s(k) \geq \rank^<$. We use the hyper-parameters: $\tau=0.01$, $\rho=100$, $\delta=0.05$.

\medbreak
We illustrate in~\cref{fig:sup_hrank_surrogate} $H^>_s$ and in~\cref{fig:sup_hrank_surrogate} $H^<_s$ \vs $s_j-s_k$. The margins denote the fact the even when the instance $k$ is correctly ranked (lower cosine similarity than $j$ in~\cref{fig:sup_hrank_surrogate} and higher in~\cref{fig:sup_hrank_surrogate}) we still want to backbropagate gradient which leads to more robust training.

\subsection{Discussion}

\textcolor{black}{HAPPIER requires the definition of the relevance function. In our work, we leverage the hierarchical tree between concepts to this end. Is this a strong assumption? We argue that the access to a hierarchical tree is not a prohibitive factor. Hierarchical trees are available for a surprising number of datasets (CUB-200-2011~\cite{WahCUB_200_2011}, Cars196~\cite{cars196}, InShop~\cite{liuLQWTcvpr16DeepFashion}, SOP~\cite{oh2016deep}), including \emph{large scale} ones (iNaturalist~\cite{van2018inaturalist}, the three DyML datasets~\cite{sun2021dynamic} and also Imagenet~\cite{imagenet}). Even when hierarchical relations are not directly available, they are not that difficult to obtain since the tree complexity depends only on the number of classes and not of examples. Hierarchical relations can be semi-automatically obtained by grouping fine-grained labels in existing datasets, as was previously done by \eg~\cite{chang2021your}. For instance, while hierarchical labels are not directly available in scene or landmarks datasets~\cite{Radenovic-CVPR18}, this could be extended to them at a reasonable cost, \eg in Paris6k ``Sacre Coeur'' might be considered closer to ``Notre Dame'' than to the ``Moulin Rouge''. The large lexical database Wordnet~\cite{wordnet} can also be used to define hierarchies between labels and define semantic similarity, as in Imagenet~\cite{imagenet} or the SUN database~\cite{SUN}. Furthermore, our approach can be extended to leverage general knowledge beyond hierarchical taxonomies, by defining more general relevance functions built on \eg continuous similarities or attributes~\cite{parikh2011relative}.}

\section{Experiments}

\subsection{Datasets}

\subsubsection{Stanford Online Product (SOP)} 
\begin{figure}[ht]
    \centering
    \includegraphics[width=0.7\textwidth]{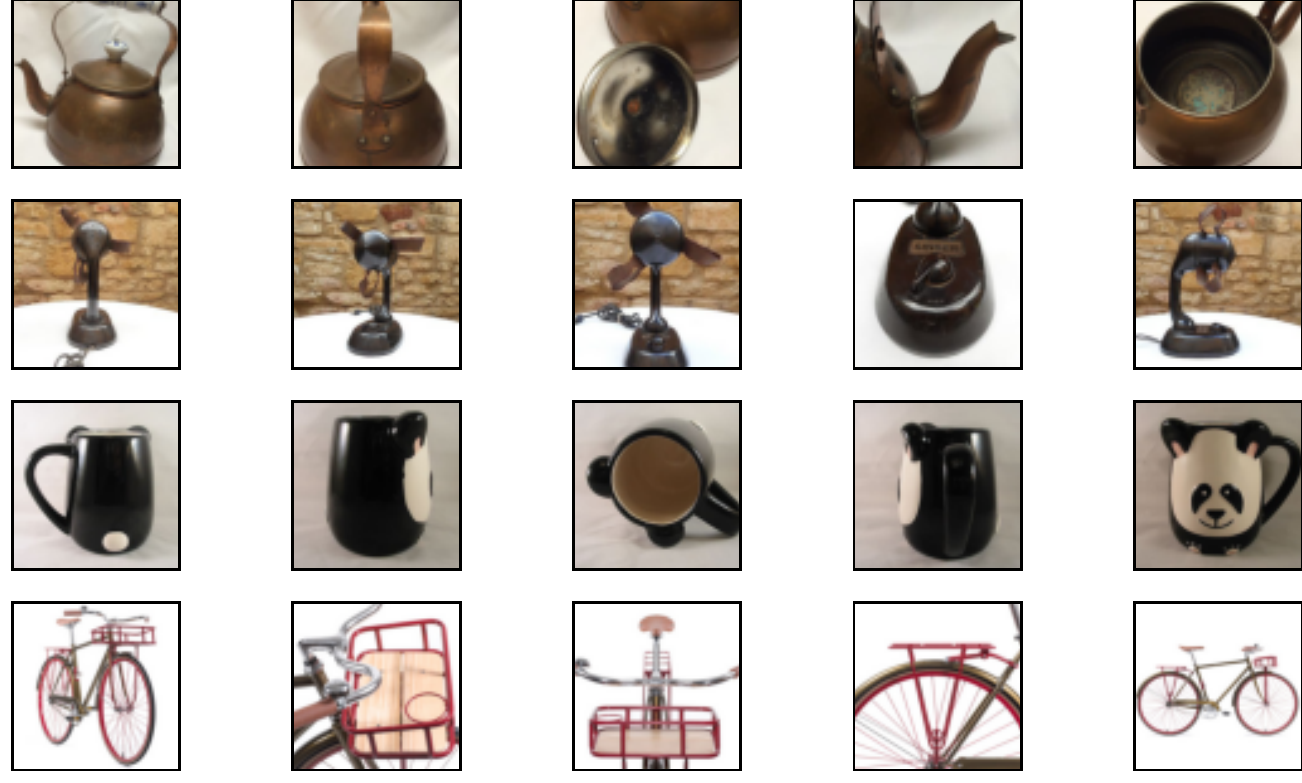}
    \caption{Images from Stanford Online Products.}
    \label{fig:sup_sop_illust}
\end{figure}

\cite{oh2016deep} is a standard dataset for Image Retrieval it has two levels of semantic scales, the object Id (fine) and the object category (coarse). It depicts Ebay online objects, with \num{120053} images of \num{22634} objects (Id) classified into \num{12} (coarse) categories (\eg bikes, coffee makers \etc), see~\cref{fig:sup_sop_illust}. We use the reference train and test splits from~\cite{oh2016deep}. The dataset can be downloaded at: \url{https://cvgl.stanford.edu/projects/lifted_struct/}.

\subsubsection{iNaturalist-2018 Base/Full} 
\begin{figure}[ht]
    \centering
    \includegraphics[width=0.7\textwidth]{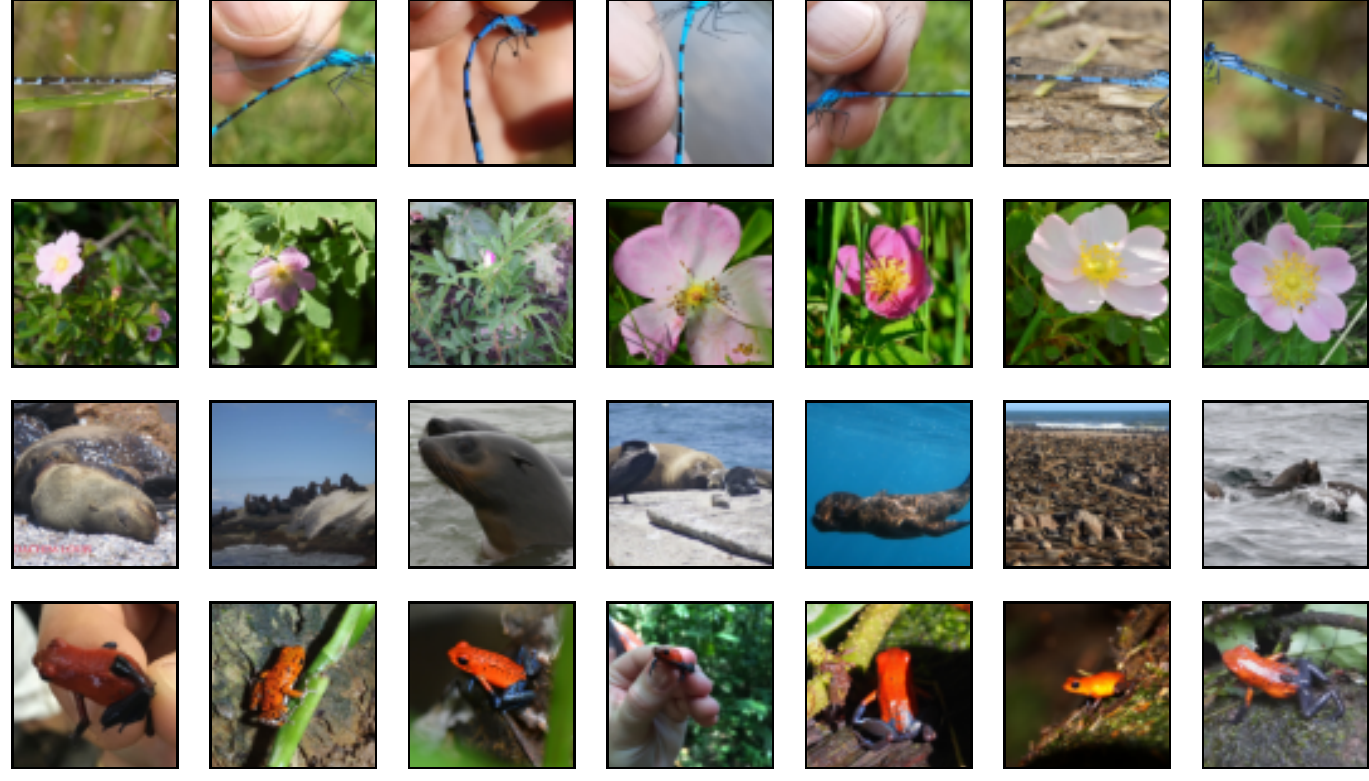}
    \caption{Images from iNaturalist-2018.}
    \label{fig:sup_inat_illust}
\end{figure}

iNaturalist-2018 is a dataset that has been used for Image Retrieval in recent works~\cite{smoothap,ramzi2021robust}. It depicts animals, plants, mushroom \etc in wildlife, see~\cref{fig:sup_inat_illust}, it has in total \num{461 939} images and \num{8142} fine-grained classes (``Species''). We use two different sets of annotations: a set of annotations with 2 semantic levels the species (fine) and intermediate scale (coarse), we term this dataset iNat-base, and the full biological taxonomy which consists of 7 semantic levels (``Species'', ``Genus'' \dots) we term this dataset iNat-full. We use the standard Image Retrieval splits from~\cite{smoothap}. The dataset can be downloaded at: \href{https://github.com/visipedia/inat_comp/tree/master/2018}{\nolinkurl{github.com/visipedia/inat_comp}}, and the retrieval splits at: \href{https://drive.google.com/file/d/1sXfkBTFDrRU3__-NUs1qBP3sf_0uMB98/view?usp=sharing}{\nolinkurl{drive.google.com}}. 

\subsubsection{DyML-datasets} The DyML benchmark~\cite{sun2021dynamic} is composed of three datasets, DyML-V that depicts vehicles, DyML-A that depicts animals, DyML-P that depicts online products. The training set has three levels of semantic ($L=3$), and each image is annotated with the label corresponding to each level (like SOP and iNat-base/full), however the test protocol is different. At test time for each dataset there is three sub-datasets, each sub-dataset aims at evaluating the model on a specific hierarchical level (\eg ``Fine''), so we can only compute binary metrics on each sub-dataset. We describe in~\cref{tab:sup_stats_dyml} the statistics of the train and test datasets. The three datasets can be downloaded at: \href{https://onedrive.live.com/?authkey=\%21AMLHa5h\%2D56ZZL94\&id=F4EF5F480284E1C2\%21106\&cid=F4EF5F480284E1C2}{\nolinkurl{onedrive.live.com}}.

\begin{table*}[ht]
    \vspace{-0.6\intextsep}
    \caption{Statistics of the three train and test DyML benchmarks~\cite{sun2021dynamic}.
    }
    \label{tab:sup_stats_dyml} 
    \centering
    \begin{tabularx}{\textwidth}{Y Y | YY | YY | YY }
        \toprule
         \multicolumn{2}{c|}{\multirow{2}{*}{Datasets}} & \multicolumn{2}{c|}{DyML-Vehicle} & \multicolumn{2}{c|}{DyML-Animal} & \multicolumn{2}{c}{DyML-Product}\\
         \cmidrule{3-8}
         && train & test & train & test & train & test \\
         \midrule
         \multirow{2}{*}{Coarse} & Classes & 5 & 6 & 5 & 5 & 36 & 6 \\\
         & Images & 343.1 K & 5.9 K & 407.8 K & 12.5 K & 747.1 K & 1.5 K \\
         \midrule
         \multirow{2}{*}{Middle} & Classes & 89 & 127 & 28 & 17 & 169 & 37 \\
         & Images & 343.1 K & 34.3 K & 407.8 K & 23.1 K & 747.1 K & 1.5 K \\
         \midrule
         \multirow{2}{*}{Fine} & Classes & 36,301 & 8,183 & 495 & 162 & 1,609 & 315 \\
         & Images & 343.1 K & 63.5 K & 407.8 K & 11.3 K & 747.1 K & 1.5 K \\
         \bottomrule
    \end{tabularx}
\end{table*}

\subsection{Implementation details}\label{sec:sup_implementation_details}

\subsubsection{SOP \& iNat-base/full} Our model is a ResNet-50~\cite{he2015deep} pretrained on Imagenet to which we append a \texttt{LayerNormalization} layer with no affine parameters after the (average) pooling and a Linear layer that reduces the embeddings size from $2048$ to $512$. We use the Adam~\cite{kingma2014adam} optimizer with a base learning rate of $1e^{-5}$ and weight decay of $1e^{-4}$ for SOP and a base learning rate of $1e^{-5}$ and weight decay of $4e^{-4}$ for iNat-base/full. The learning rate is decreased using cosine annealing decay, for 75 epochs on SOP and 100 epochs on iNat-base/full. We ``warm up'' our model for 5 epochs, \ie the pretrained weights are not optimized. We use standard data augmentation: \texttt{RandomResizedCrop} and \texttt{RandomHorizontalFlip}, with a final crop size of $224$, at test time we use \texttt{CenterCrop}. We set the random seed to $0$ in all our experiments. We use a fixed batch size of 256 and use the hard sampling strategy from~\cite{cakir2019deep} on SOP and the standard class balanced sampling~\cite{zhai2018classification} (4 instances per class) on iNat-base/full.

\subsubsection{DyML} We use a ResNet-34~\cite{he2015deep} randomly initialized on DyML-V\&A and pretrained on Imagenet for DyML-P, following~\cite{sun2021dynamic}. We use an SGD optimizer with Nesterov momentum (0.9), a base learning rate of $0.1$ on DyML-V\&A and $0.01$ on DyML-P with a weight decay of $1e^{-4}$. We use cosine annealing decay to reduce the learning rate for $100$ epochs on DyML-V\&A and $20$ on DyML-P. We use the same data augmentation and random seed as for SOP and iNat-base. We also use the class balanced sampling (4 instances per class) with a fixed batch size of $256$.

\subsection{Metrics}

The ASI~\cite{fagin2003comparing} measures at each rank $n\leq N$ the set intersection proportion ($SI$) between the ranked list $a_1,\dots,a_N$ and the ground truth ranking $b_1,\dots,b_N$, with $N$ the total number of positives. As it compares intersection the ASI can naturally take into account the different levels of semantic: 

\begin{align*}
    SI(n) &= \frac{|\{a_1,\dots,a_n\}\cap\{b_1,\dots,b_n\}|}{n} \\
    ASI &= \frac{1}{N} \sum_{n=1}^N SI(n) 
\end{align*}

The NDCG~\cite{croft2010search} is the reference metric in information retrieval, we define it using the semantic level $l$ of each instance:

\begin{align*}
    &DCG = \sum_{k\in\Omega^+} \frac{2^l - 1}{\log_2(1+\rank(k))}, \; \text{with $k\in\Omega^{(l)}$.} \\
    &NDCG = \frac{DCG}{\max_{\text{ranking}} DCG}
\end{align*}

To compute the AP for the semantic level $l$ we consider that all instances with semantic levels $\geq l$ are positives:
\begin{equation*}
    AP^{(l)} = \sum_{k\in\bigcup_{q=l}^L \Omega^{(q)}} \frac{\rank^l(k)}{\rank(k)}, \; \text{where } \rank^l(k) = 1 + \sum_{j\in\bigcup_{q=l}^L \Omega^{(q)}} H(s_j-s_k)
\end{equation*}

\subsection{Source Code}

Our code is based on \texttt{PyTorch}~\cite{pytorch}. We use utilities from \texttt{Pytorch Metric Learning}~\cite{PML} \eg for samplers and losses, \texttt{Hydra}~\cite{hydra} to handle configuration files (Yaml), \texttt{tsnecuda}~\cite{chan2019gpu} to compute t-SNE reductions using GPUs and standard Python libraries such as \texttt{NumPy}~\cite{harris2020array} or \texttt{Matplotlib}~\cite{matplotlib}.

We use the publicly available implementations of the NSM loss~\cite{zhai2018classification}\footnote{\url{https://github.com/azgo14/classification_metric_learning}} which is under an Apache-2.0 license, of NCA++\cite{teh2020proxynca++}\footnote{\url{https://github.com/euwern/proxynca_pp}} which is under an MIT license, of ROADMAP~\cite{ramzi2021robust}\footnote{\url{https://github.com/elias-ramzi/ROADMAP}} which is under an MIT license, we use the implementation of \texttt{Pytorch Metric Learning}~\cite{PML}\footnote{\url{https://github.com/KevinMusgrave/pytorch-metric-learning}} for the $\text{TL}_{\text{SH}}$~\cite{wu2017sampling} (MIT license), and finally we have implemented the CSL~\cite{sun2021dynamic} after discussion with the authors and we will make it part of our repository.

We had access to both Nvidia Quadro RTX 5000 and Tesla V-100 (16 GiB GPUs). We use mixed precision training~\cite{micikevicius2017mixed}, which is native to \texttt{PyTorch}, to accelerate training, making our models train for up to 7 hours on Stanford Online Products, 25 hours on iNaturalist-2018, less than 20 hours on both DyML-A and DyML-V and 6 hours on DyML-P.

\subsection{On DyML results} Their is no public code available to reproduce the results of~\cite{sun2021dynamic}. After personal correspondence with the authors, we have been able to re-implement the CSL method from~\cite{sun2021dynamic}. We report the differences in performances between our results and theirs in~\cref{tab:sup_discrepancies}. Our implementation of CSL performs better on the three datasets which is the results of our better training recipes detailed in~\cref{sec:sup_implementation_details}. Our discussions with the authors of~\cite{sun2021dynamic} confirmed that the performances obtained with our re-implementation of CSL are valid and representative of the method’s potential.

\begin{table*}[ht]
    \caption{Difference in performances for CSL between results reported in~\cite{sun2021dynamic} and our experiments on the DyML benchmarks.
    }
    \label{tab:sup_discrepancies} 
    \centering
    \begin{tabularx}{\textwidth}{l l YYY | YYY | YYY }
        \toprule
         & \multirow{2}{*}{Method} & \multicolumn{3}{c|}{DyML-Vehicle} & \multicolumn{3}{c|}{DyML-Animal} & \multicolumn{3}{c}{DyML-Product}\\
        \cmidrule{3-11}
         && mAP & ASI & R@1 & mAP & ASI & R@1 & mAP & ASI & R@1 \\
         \midrule
         & CSL~\cite{sun2021dynamic} & 12.1 & 23.0 & 25.2 & 31.0 & 45.2 & 52.3 & 28.7 & 29.0 & 54.3 \\
          & CSL (ours) & 30.0 & 43.6 & 87.1 & 40.8 & 46.3 & 60.9 & 31.1 & 40.7 & 52.7 \\
         \bottomrule
    \end{tabularx}
\end{table*}

\section{Qualitative results}

\subsection{Robustness to $\lambda$}

\begin{figure}[ht]
    \centering
        
    \begin{subfigure}[ht]{0.3\textwidth}
        \includegraphics[width=1\textwidth]{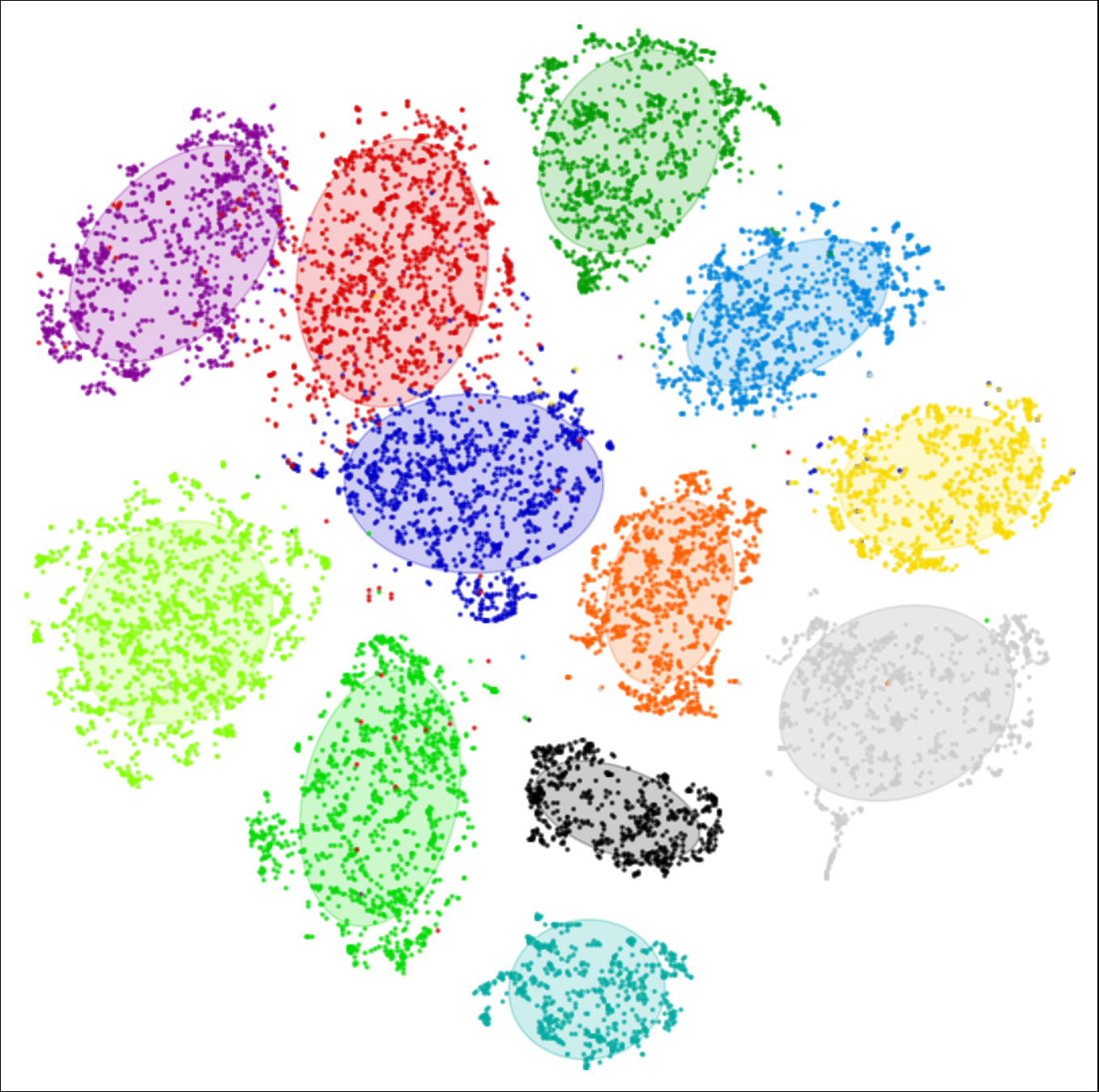}
        \caption{$\lambda=0.3$}
        \label{fig:tsne_happier_03}
    \end{subfigure}
    ~
    \begin{subfigure}[ht]{0.3\textwidth}
    \includegraphics[width=1\textwidth]{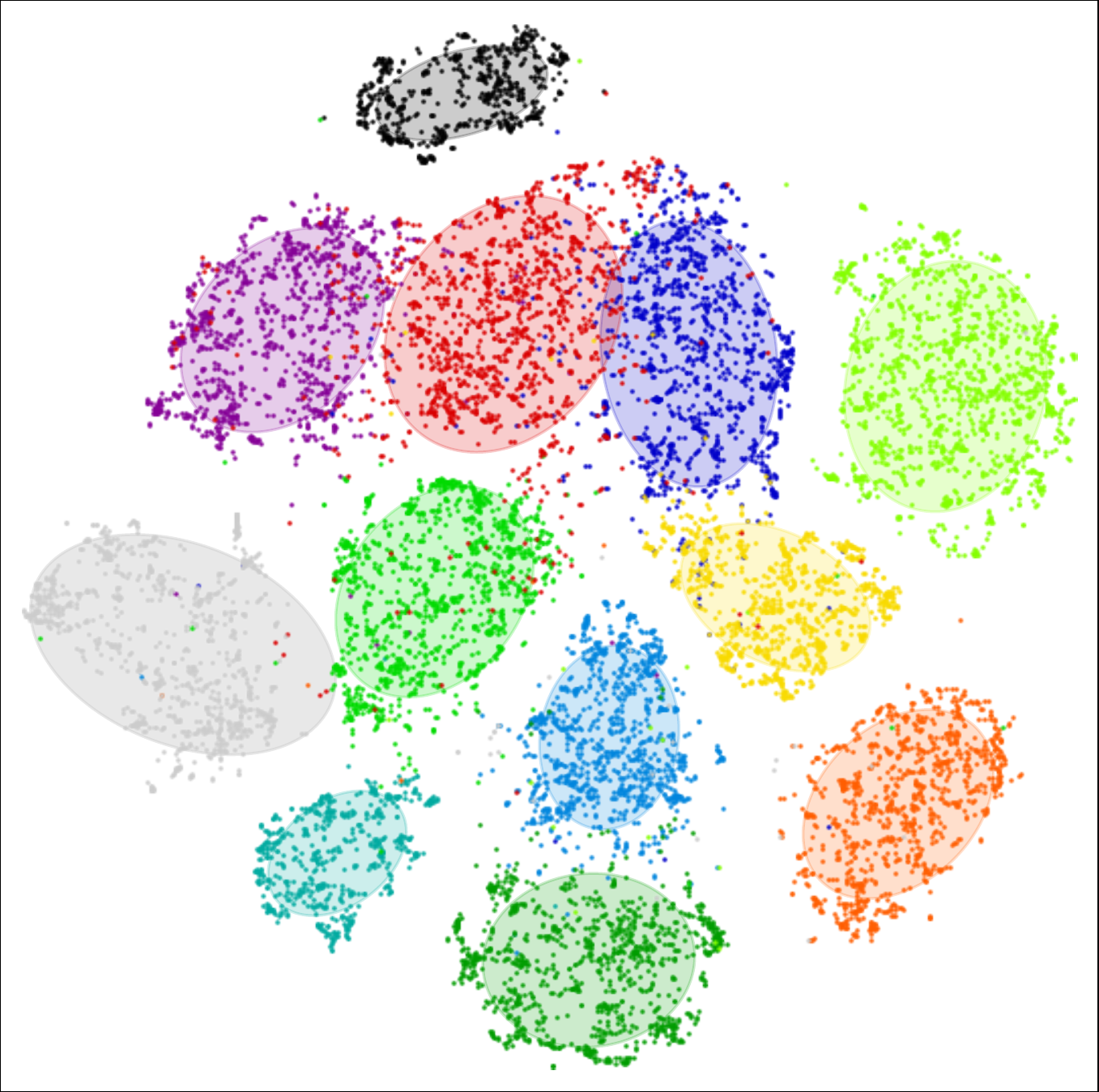}
    \caption{$\lambda=0.5$}
    \label{fig:tsne_happier_05}
    \end{subfigure}%
    ~
    \begin{subfigure}[ht]{0.3\textwidth}
    \includegraphics[width=1\textwidth]{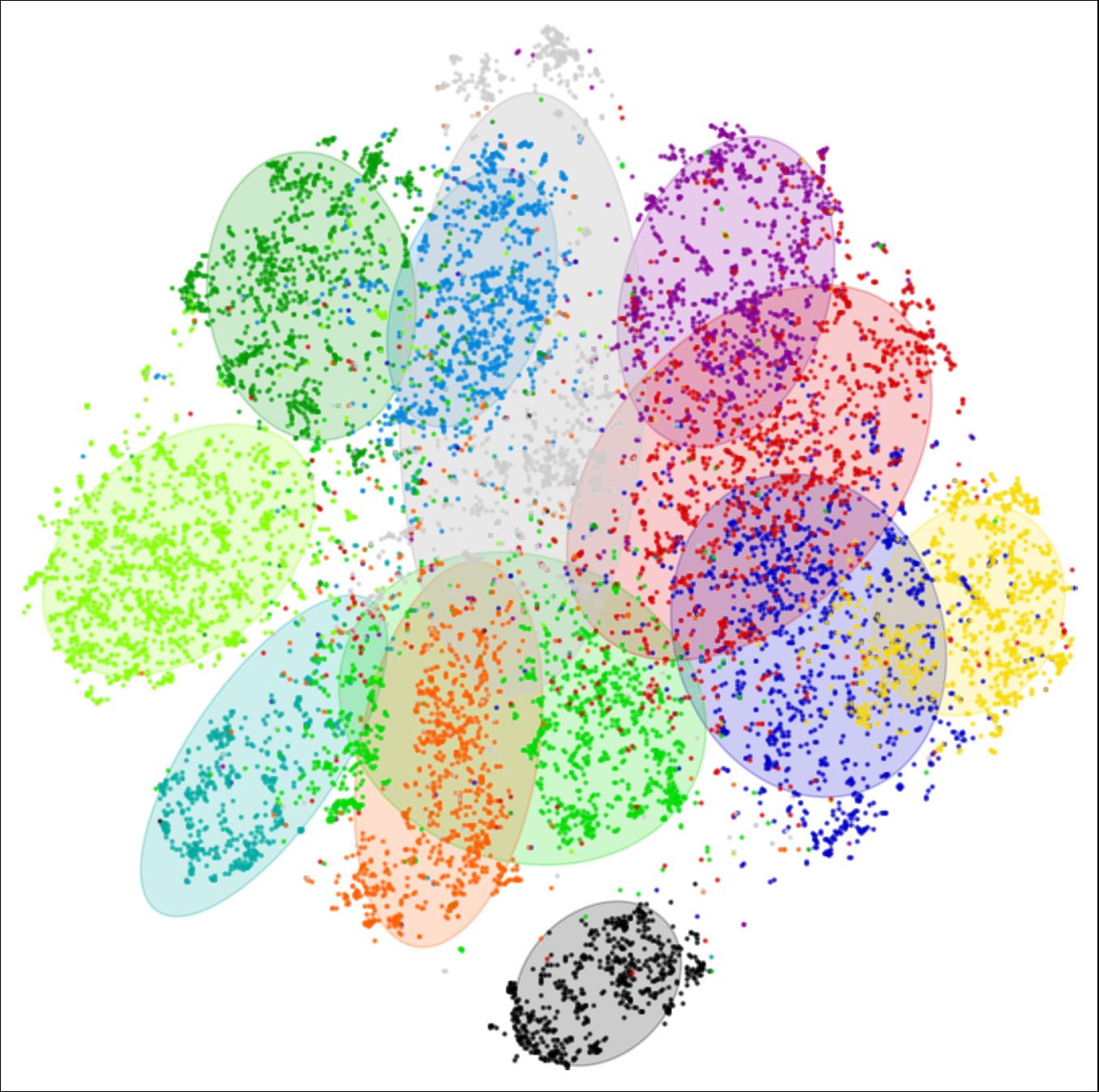}
    \caption{$\lambda=0.9$}
    \label{fig:tsne_happier_09}
    \end{subfigure}%

    \caption{t-SNE visualisation of the embedding space of models trained with HAPPIER on SOP with different values of $\lambda$. Each point is the average embedding of each fine-grained label (object instance) and the colors represent coarse labels (object category, \eg bike, coffee maker).}
    \label{fig:tsne_sup}
\end{figure}

\textcolor{black}{Fig. 4b of the main paper illustrates that HAPPIER is robust with respect to $\lambda$ with performances increasing for most values between 0.1 and 0.9. In addition, we also show in~\cref{fig:tsne_sup} that for $0<\lambda<0.9$ HAPPIER leads to a better organization of the embedding space than a fine-grained baseline (see Fig. 4a in main paper). This is expected since the lower $\lambda$ is, the more emphasis is put on optimizing $\hap$, which organizes the embedding space in a hierarchical structure.}

\subsection{Comparison of HAPPIER \vs CSL}

\begin{figure}[ht]
    \centering
        
    \begin{subfigure}[t]{\textwidth}
        \includegraphics[width=\textwidth]{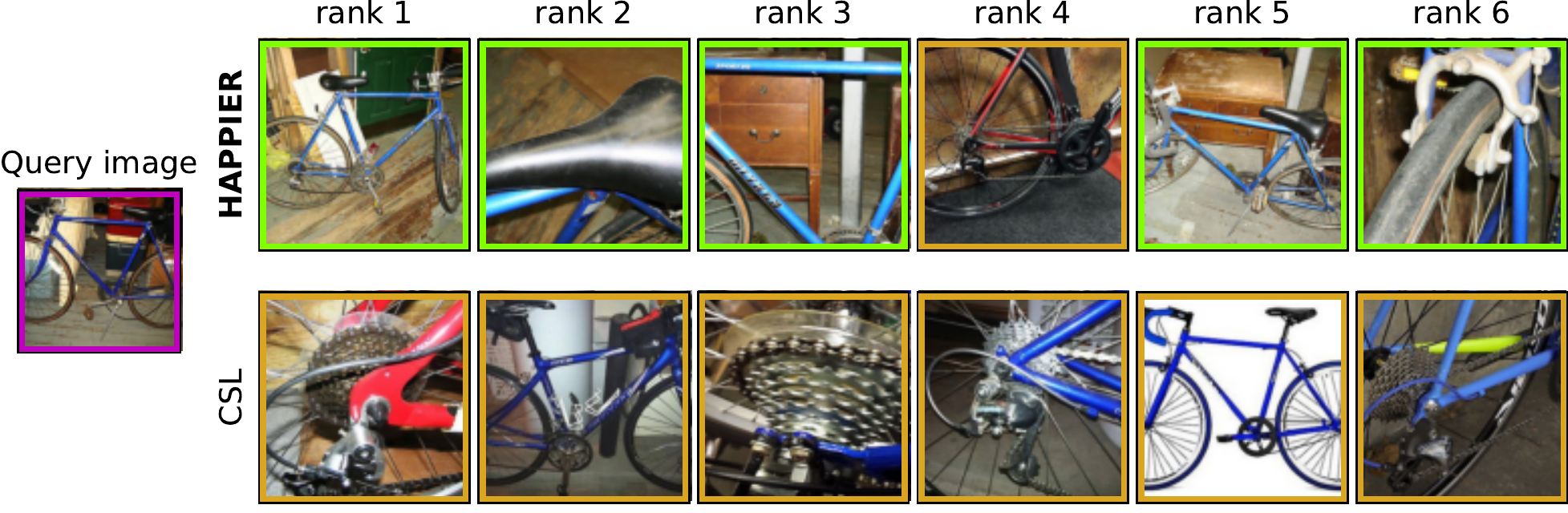}
    \end{subfigure}
    
    \begin{subfigure}[t]{\textwidth}
    \includegraphics[width=\textwidth]{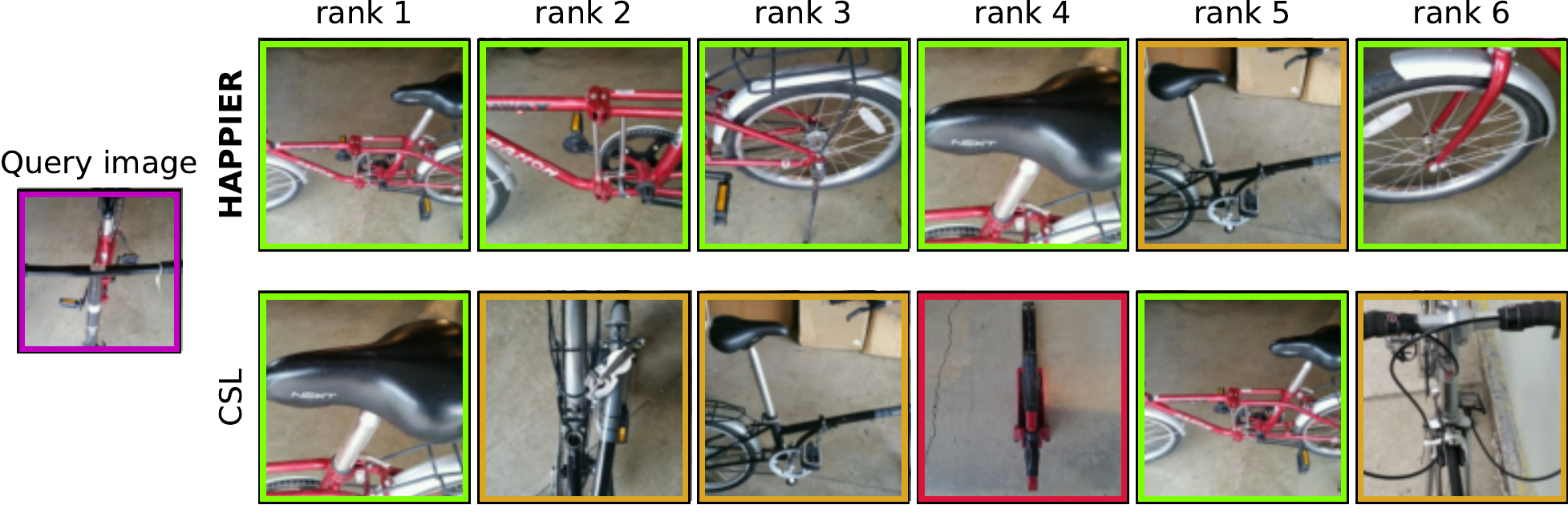}
    \end{subfigure}%

    \caption{Qualitative comparison of \HAPPIER \vs CSL~\cite{sun2021dynamic} on SOP}
    \label{fig:happier_vs_csl}
\end{figure}

\textcolor{black}{In~\cref{fig:happier_vs_csl}, we compare HAPPIER against the hierarchical image retrieval method CSL~\cite{sun2021dynamic}. We observe qualitative improvements where HAPPIER results in a better ranking. This highlights the benefit of optimizing directly a hierarchical metric, \ie $\hap$, rather than optimizing a proxy based triplet loss as in CSL.}

\subsection{Controlled errors: iNat-base} 

\begin{figure}[ht]
    \centering
        
    \begin{subfigure}[t]{\textwidth}
        \includegraphics[width=1\textwidth]{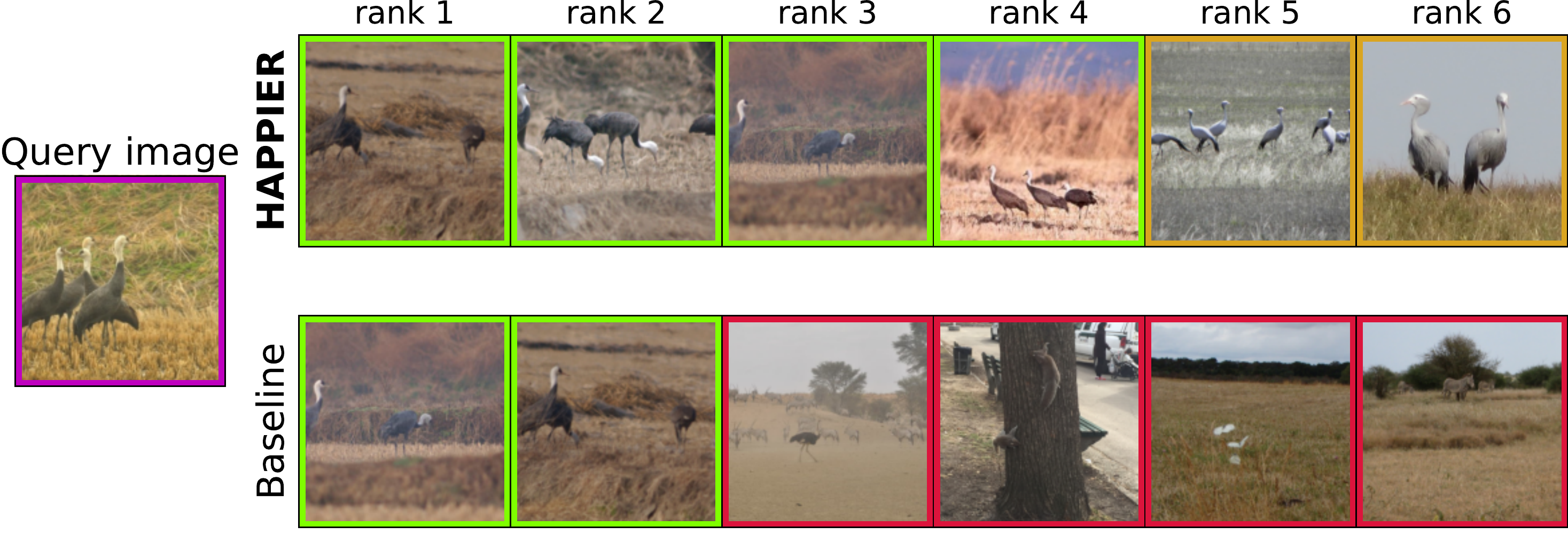}
        \caption{\HAPPIER can help make less severe mistakes. The inversion on the bottom row are with negative instances (in \textcolor{red}{red}), where as with \HAPPIER (top row) inversions are with instances sharing the same coarse label (in \textcolor{orange}{orange}).}
        \label{fig:sup_qual_inat_good}
    \end{subfigure}
    
    \begin{subfigure}[t]{\textwidth}
    \includegraphics[width=1\textwidth]{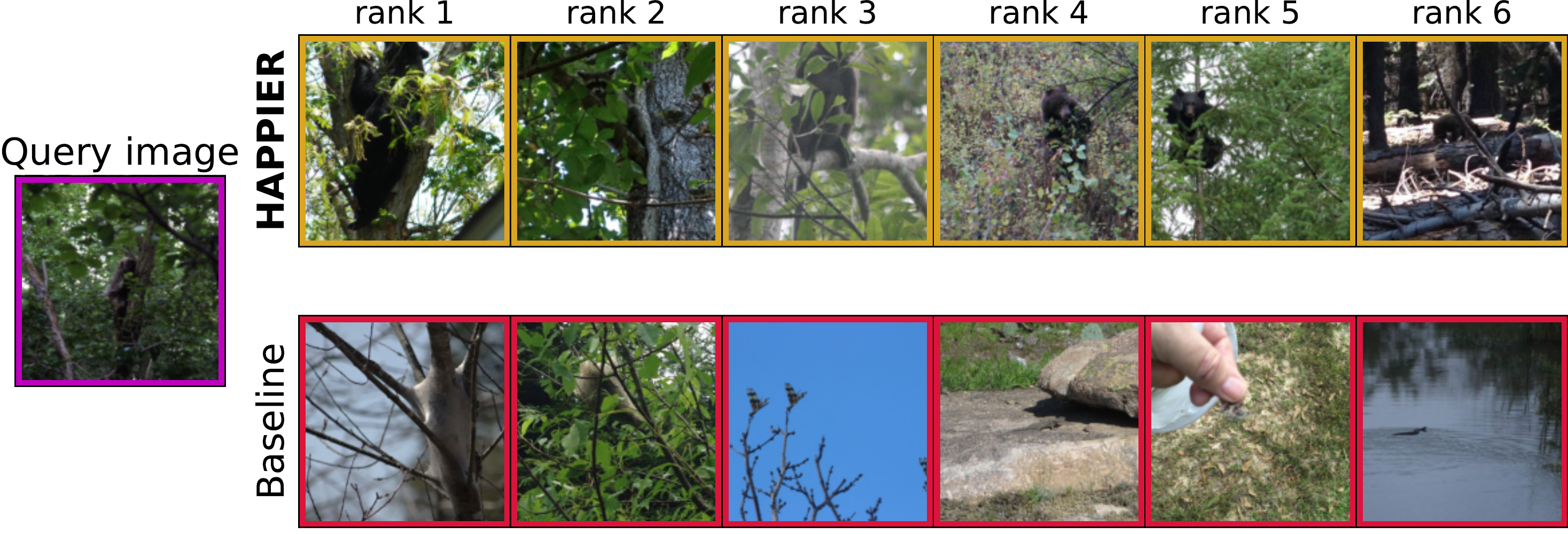}
    \caption{In this example, the models fail to retrieve the correct fine grained images. However \HAPPIER still retrieves images with the same coarse label (in \textcolor{orange}{orange}) whereas the baseline retrieves images that are dissimilar semantically to the query (in \textcolor{red}{red}).}
    \label{fig:sup_qual_inat_error}
    \end{subfigure}%

    \caption{Qualitative examples of failure cases from a standard fine-grained model corrected by training with \HAPPIER.}
    \label{fig:sup_qualitative_results}
\end{figure}

We showcase in~\cref{fig:sup_qualitative_results} errors of \HAPPIER \vs a fine-grained baseline on iNat-base. On~\cref{fig:sup_qual_inat_good}, we illustrate how a model trained with \HAPPIER makes mistakes that are less severe than a baseline model trained only on the fine-grained level. On~\cref{fig:sup_qual_inat_error}, we show an example where both models fail to retrieve the correct fine-grained instances, however the model trained with \HAPPIER retrieves images of bikes that are semantically more similar to the query.

\subsection{Controlled errors: iNat-full} 

\begin{figure}[ht]
    \centering
    \includegraphics[width=1\textwidth]{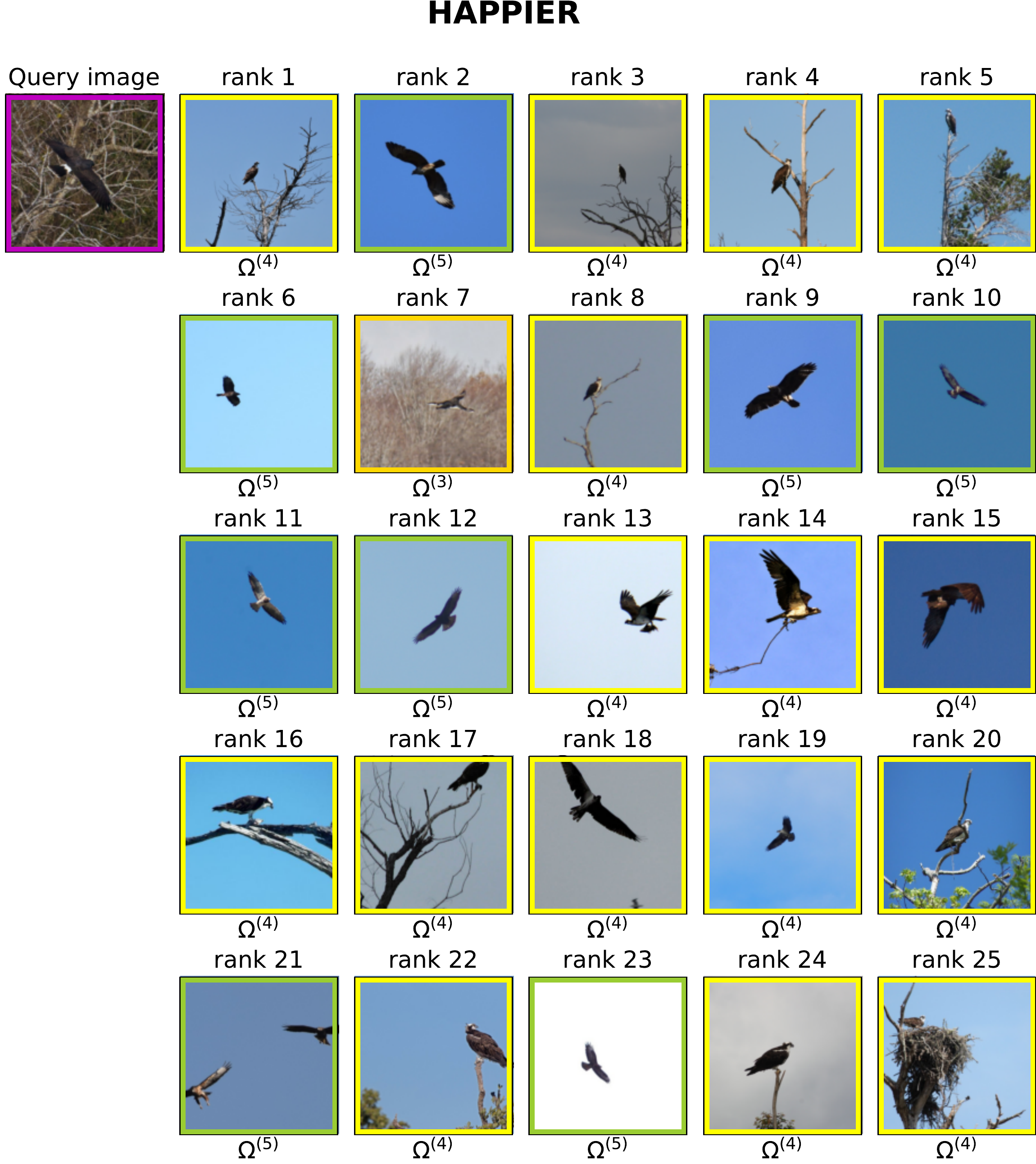}
    \caption{Images retrieved for the \textcolor{amethyst}{query image} by a model trained with \textbf{\HAPPIER} on iNat-full ($L=7$).}
    \label{fig:sup_qual_inat_full_happier}
\end{figure}

\begin{figure}[ht]
    \centering
    \includegraphics[width=1\textwidth]{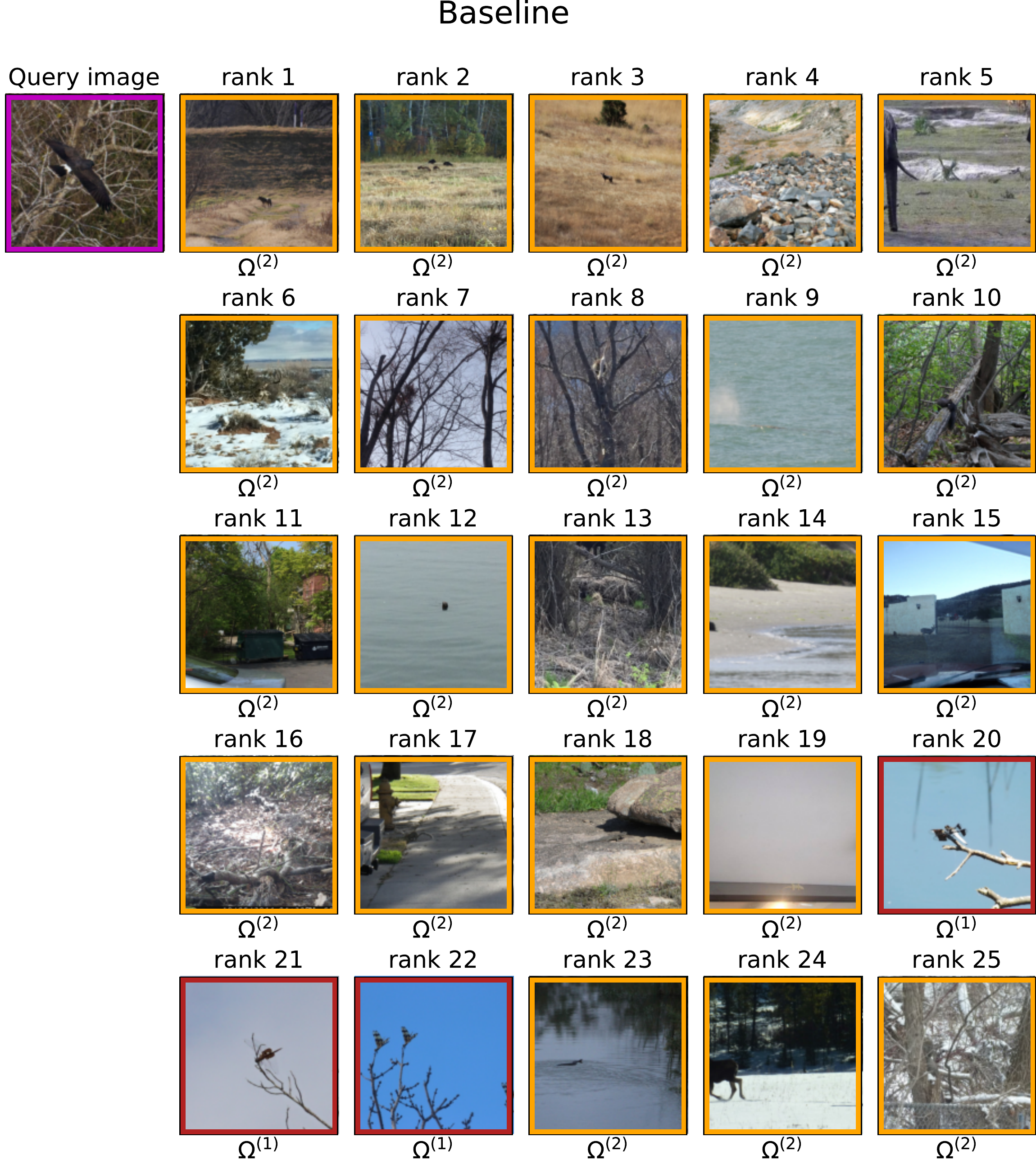}
    \caption{Images retrieved for the \textcolor{amethyst}{query image} by a model trained with standard model on iNat-full ($L=7$).}
    \label{fig:sup_qual_inat_full_baseline}
\end{figure}

We illustrate in~\cref{fig:sup_qual_inat_full_happier,fig:sup_qual_inat_full_baseline} an example of a query image and the top $25$ retrieved results on iNat-full ($L=7$). Given the same query both models failed to retrieve the correct fine-grained images (that would be in $\Omega^{(7)}$). The standard model in~\cref{fig:sup_qual_inat_full_baseline} retrieves images that are semantically more distant than the images retrieved with \HAPPIER in~\cref{fig:sup_qual_inat_full_happier}. For example \HAPPIER retrieves images that are either in $\Omega^{(5)}$ or $\Omega^{(4)}$ (only one instance is in $\Omega^{(3)}$) whereas the standard model retrieves instances that are in $\Omega^{(2)}$ or $\Omega^{(1)}$.

